\documentclass{article} 
\PassOptionsToPackage{table}{xcolor}
\usepackage{iclr2027_conference,times}
\usepackage[table]{xcolor}

\usepackage{amsmath,amsfonts,bm}

\def\eqref#1{equation~\ref{#1}}

\def\1{\bm{1}}

\DeclareMathAlphabet{\mathsfit}{\encodingdefault}{\sfdefault}{m}{sl}
\SetMathAlphabet{\mathsfit}{bold}{\encodingdefault}{\sfdefault}{bx}{n}

\usepackage{hyperref}
\usepackage{url}

\usepackage{graphicx} 
\usepackage{soul} 

\usepackage{booktabs}
\usepackage{graphicx}
\usepackage{multirow}
\usepackage{makecell}
\usepackage{enumitem}
\usepackage{float}

\usepackage[capitalise]{cleveref}
\crefname{equation}{Eq.}{Eqs.}
\creflabelformat{equation}{(#2#1#3)}
\renewcommand{\eqref}[1]{\cref{#1}}

\newcommand{\hdrspan}[1]{%
  \multirow{2}{*}[-0.5\dimexpr\aboverulesep+\belowrulesep+\cmidrulewidth\relax]%
    {\makecell{#1}}}

\title{SpecKCast: Spectral Kalman Forecasting for Diffusion Acceleration}

\author{Javad Rajabi$^{1,2,3}$\thanks{Work done during an internship at Samsung.}\hspace{0.45em}, Hakki Can Karaimer$^{3}$, David B.~Lindell$^{1,2}$, \\
\textbf{Babak Taati$^{1,2}$, Michael S.~Brown$^{3}$} \\
$^1$University of Toronto \quad $^2$Vector Institute \\
$^3$AI Center--Toronto, Samsung Electronics \\
\texttt{rajabi@cs.toronto.edu}
}

\iclrfinalcopy 
\begin{document}

\maketitle
\lhead{Preprint. Under review.}


\vspace{-15pt}

\begin{abstract}
\vspace{-3pt}
Diffusion Transformers (DiTs) achieve high-fidelity image and video generation, but they suffer from substantial inference costs due to repeated transformer evaluations across many denoising steps. Feature caching reduces this cost by reusing features across denoising steps. Recent forecasting methods extend this idea by extrapolating features to skipped steps instead of reusing them. However, existing forecasting methods extrapolate the features they predict uniformly, even though the spatial-frequency components of these features change most at different stages of denoising. They also overlook that denoising progresses nonuniformly, so equal step intervals can correspond to very different amounts of change. These mismatches amplify forecast errors as more steps are skipped, limiting the achievable acceleration. We propose SpecKCast, a training-free method that forecasts the model output in the frequency domain. SpecKCast applies a 2D discrete cosine transform (DCT) to the output and tracks each coefficient with a lightweight Kalman filter along a band-specific time coordinate derived from the noise schedule. The filter tracks each coefficient's value and rate of change, and the transformer is evaluated only when the filter's forecast uncertainty exceeds a threshold. Experiments on FLUX.1-dev, Qwen-Image, Wan2.1-14B, and HunyuanVideo show that SpecKCast achieves a better trade-off between acceleration and fidelity than existing caching and forecasting methods, reaching up to $6\times$ speedup with less than 0.05\,GiB of additional memory.
\end{abstract}

\vspace{-5pt}
\begin{figure}[H]
  \begin{center}
    \includegraphics[width=0.95\textwidth]{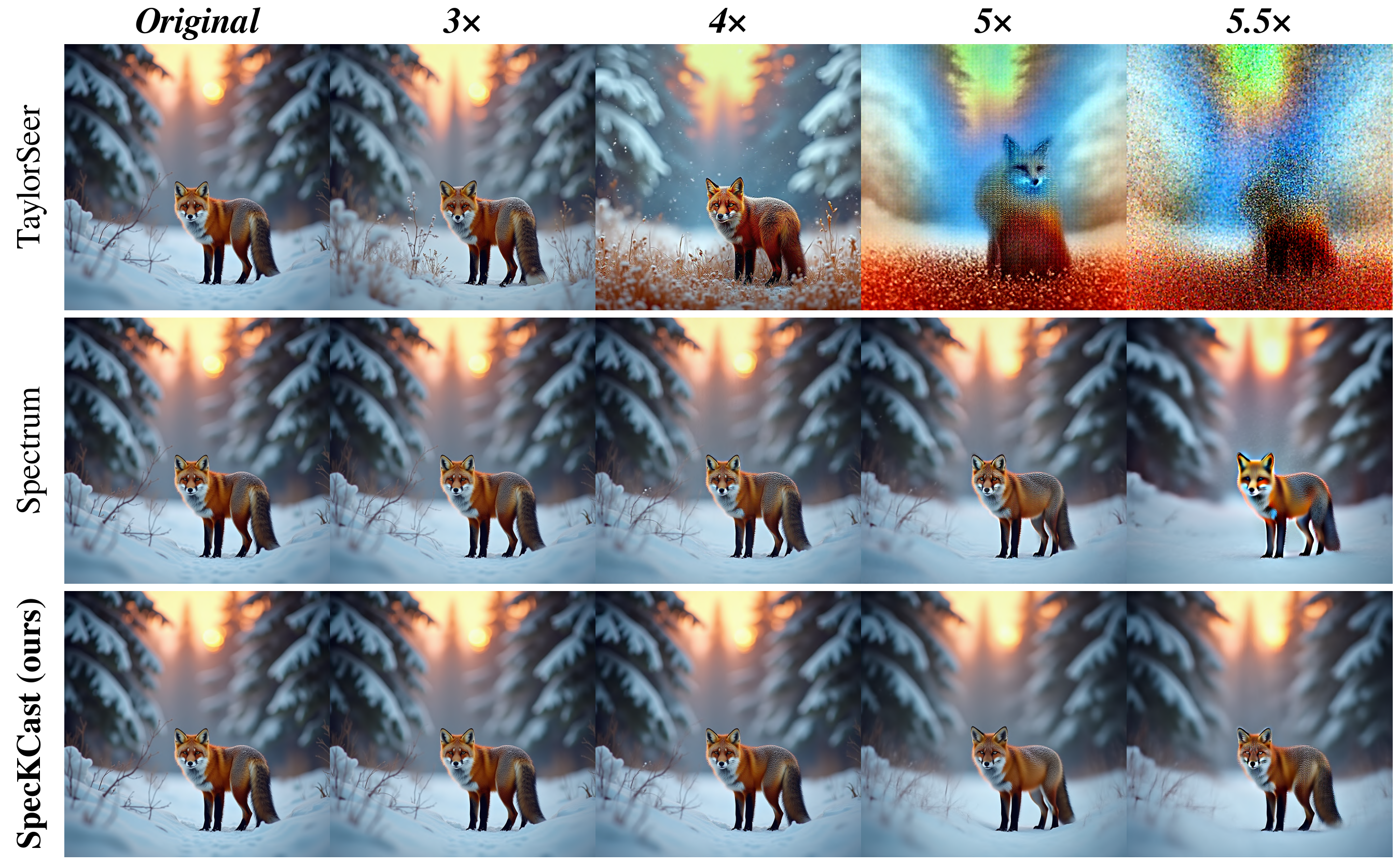}
  \end{center}
  \vspace{-13pt}
  \caption{\textbf{Qualitative comparison} on FLUX at different acceleration ratios.}
  \label{fig:teaser}
\end{figure}


\vspace{-5pt}
\section{Introduction}
\label{sec:intro}
\vspace{-5pt}

Diffusion models  \citep{ho2020denoising} have become the dominant paradigm for high-fidelity visual generation, gaining significant attention from recent advances in image and video generation. Diffusion Transformers (DiTs) \citep{peebles2023scalable} further improved generation quality by enabling greater model capacity. However, their costly forward passes throughout many denoising steps lead to substantially higher inference latency, limiting their use in real-time applications. To address this bottleneck, a wide range of acceleration techniques have been proposed to reduce either the step count or the per-step cost, including efficient attention \citep{xia2025training}, step distillation \citep{salimans2022progressive}, and quantization \citep{li2023q}.

Among these, feature caching has emerged as a practical, training-free approach. Motivated by the observation that model outputs are highly similar across adjacent timesteps, these methods cache features from previous steps and reuse them in subsequent ones, skipping substantial computation \citep{deepcache,fora}. However, this similarity decays as the gap between timesteps grows, so naive reuse over longer spans significantly degrades generation quality, limiting caching to modest acceleration ratios. To address this, recent work shifts from reusing features to forecasting them \citep{taylorseer, spectrum}. Rather than assuming that a cached feature remains valid at a later timestep, forecasting methods exploit temporal continuity along the denoising trajectory, using features from previous steps to estimate their evolution and extrapolate their values at the target timestep. By modeling temporal drift, forecasting remains effective across wider timestep gaps, enabling higher acceleration ratios at comparable generation quality.

Despite their promise, existing forecasting approaches impose uniformity on both what they forecast and how they model the signal over time \citep{taylorseer, spectrum}. First, they extrapolate the model output or feature representation as a whole, or apply the same forecasting rule uniformly to all of its components (e.g., channels), even though the spatial-frequency components of the signal evolve very differently. As \Cref{fig:fig1_velocity} shows, low-frequency content changes sharply in the first few steps and then settles, while high-frequency content changes slightly at first but evolves rapidly toward the end. Because these bands are active at different stages, a single uniform forecaster may not serve them all. Second, existing methods use the sampler’s step index or raw diffusion timestep as their temporal coordinate, even though the amount of denoising performed at each step is nonuniform and depends on the scheduler. Therefore, equal intervals in the chosen coordinate can represent very different amounts of denoising progress. Treating these intervals as equivalent makes the resulting trajectory harder to model and forecast. Together, these limitations restrict achievable acceleration, amplify forecast errors, and degrade generation quality as the skip length increases.

Motivated by this insight, we propose SpecKCast, a training-free method that rethinks both what we forecast and how we extrapolate. The key idea is to forecast the model's temporal evolution separately in the frequency domain, rather than uniformly in feature or latent space. SpecKCast transforms the model output into the spatial-frequency domain with a 2D discrete cosine transform (DCT), and forecasts each coefficient during denoising. Moreover, instead of indexing the trajectory by the sampler’s timestep, SpecKCast exploits a time coordinate derived from the noise scheduler and applies a nonlinear time warping that improves the accuracy of local extrapolation within each band. To track each coefficient as it moves along its trajectory, SpecKCast uses a lightweight Kalman filter, a recursive state-space estimator that maintains the coefficient’s value and rate of change, projecting this state across a skip and correcting it after a full denoising step. The filter also estimates the uncertainty of each forecast. As more steps are skipped, this uncertainty increases, while the rate of change decays to keep the prediction stable. Furthermore, SpecKCast partitions the coefficients into frequency bands that share a common temporal coordinate and forecast uncertainty, while each coefficient keeps its own value and rate of change. Finally, at skipped steps, the predicted coefficients are transformed back to the spatial domain and passed directly to the sampler.

The main contributions of this work are summarized as follows.
\vspace{-7pt}
{\setlength{\leftmargini}{2.5em}
\begin{itemize}
    \item \textbf{Frequency-Decoupled and Scheduler-Aligned Forecasting.}
    We introduce SpecKCast, a training-free method that decomposes the model output in the spatial-frequency domain and forecasts its components along a nonlinear time coordinate derived from the noise scheduler.

    \item \textbf{Uncertainty-Aware Kalman Forecasting.}
    SpecKCast uses a lightweight Kalman filter to track frequency coefficient trajectories and leverage forecast uncertainty to maintain stability over skipped denoising steps.

    \item \textbf{State-of-the-Art Performance.}
    SpecKCast consistently improves the acceleration and generation quality trade-off over existing caching and forecasting methods while introducing negligible computational and memory overhead, requiring about 5 MiB of additional memory for FLUX.1 \citep{flux}.
\end{itemize}




\begin{figure}[t]
  \centering
  \includegraphics[width=\textwidth]{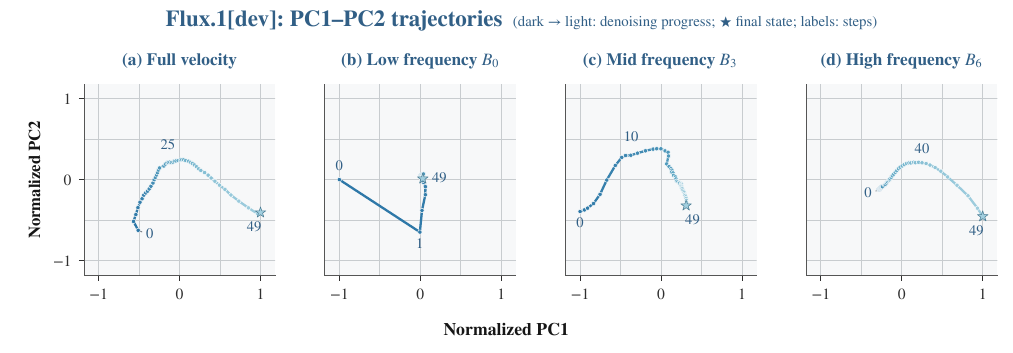}
  \vspace{-15pt}
  \caption{\textbf{Spatial-frequency bands of the denoising velocity follow different trajectories.} PCA trajectories of the 50 velocity predictions from a single FLUX.1-dev generation, projected onto the first two principal components. Larger gaps between consecutive points indicate larger changes. \textbf{(a)} Sampled entries of the full velocity. \textbf{(b--d)} Radial DCT bands: $B_0$ moves most in the first step, $B_3$ changes over a broader interval, and $B_6$ moves most after step 40. Each panel has its own PCA basis and normalization. Additional views, prompts, and Qwen-Image results are in \Cref{app:trajectories}.}
  \label{fig:fig1_velocity}
  \vspace{-15pt}
\end{figure}


\vspace{-10pt}
\section{Related work}
\label{sec:relatedwork}
\vspace{-10pt}

Diffusion models \citep{ho2020denoising} have become a leading approach for image and video generation. Since the introduction of the Diffusion Transformer (DiT) \citep{peebles2023scalable}, many works have built on this architecture, achieving strong results across diverse domains \citep{pixart, cogvideox, sdmatte, sega, proxypose}. This progress, however, comes at a cost. Sampling requires many sequential forward passes through a large network, making inference slow and expensive. This limits practical applications and has made diffusion model acceleration an important research direction. A large body of work has therefore focused on reducing sampling steps, mostly by distilling a multi-step teacher into a few-step student \citep{meng2023distillation}. Examples include consistency models \citep{song2024improved, song2023consistency} and DMD \citep{dmd, yin2024one}. Although effective, these approaches require additional training and are prone to reduced sample diversity. 

A more recent line of work explores feature caching, which accelerates inference without any additional training. The core idea is to reuse intermediate features across timesteps to avoid redundant computation \citep{toca, meancache, eigencache, ertacache}. Early methods such as DeepCache \citep{deepcache} accelerated the U-Net backbone using handcrafted caching rules, and $\Delta$-DiT \citep{delta} later extended this idea to DiT architectures. Subsequent methods further refine what and when to cache. FORA \citep{fora} reuses attention and MLP outputs across timesteps, while TeaCache \citep{teacache} exploits the correlation between timestep embeddings and model outputs to estimate how much the output changes between steps and decide when to reuse cached features. HiCache \citep{hicache} forecasts features with scaled Hermite polynomials to capture their non-monotonic dynamics in DiTs. SeaCache \citep{seacache} instead improves the caching decision by measuring feature changes after a spectral filter that preserves content-relevant components and suppresses noise. However, feature similarity decreases as the timestep gap increases. As more steps are skipped, the cached features deviate further from the actual ones, and these errors accumulate during sampling. This causes the trajectory to drift and degrades generation quality, limiting the achievable speedup.

To address this, recent work shifts from reusing features to forecasting them \citep{taylorseer, spectrum}. Instead of assuming that cached features remain valid at later timesteps, forecasting methods use features from previous steps to predict how they evolve and estimate their values at the target timestep. TaylorSeer \citep{taylorseer} does this locally with a Taylor expansion, using finite differences of the most recent cached features, while Spectrum \citep{spectrum} fits a Chebyshev expansion over the full feature history to better capture long-range dynamics. By modeling how features evolve, these methods remain effective over wider timestep gaps, enabling higher acceleration at comparable quality. However, both forecast uniformly in feature space and along the sampler's timestep, ignoring that denoising progress is nonuniform across steps. In contrast, SpecKCast forecasts each frequency band separately along a scheduler-derived time axis, using a lightweight Kalman filter that adapts its forecasts as the trajectory evolves.




\section{Background}
\label{sec:background}
\vspace{-5pt}

\subsection{Diffusion Models and Flow Matching}
\label{sec:bg_diffusion}
\label{sec:background}
\vspace{-5pt}

Diffusion models \citep{ho2020denoising} define a generative process that gradually transforms samples from a simple prior (e.g., Gaussian) into data via a learned reverse-time process. Flow matching \citep{lipman2022flow,liu2022flow} instead learns a velocity field that transports noise to data along near-straight paths. Both are instances of stochastic interpolants \citep{albergo2025stochastic}, with $x_t = \alpha_t x_0 + \sigma_t \epsilon$, where rectified flow sets $\alpha_t = 1-t$ and $\sigma_t = t$ and learns the velocity $v = \epsilon - x_0$. Sampling integrates the reverse flow over decreasing noise levels using Euler updates:
\begin{equation}
x_{j+1}
=
x_j+(\sigma_{j+1}-\sigma_j)Y_j,
\qquad
Y_j=v_\theta(x_j,\sigma_j,c),
\label{eq:euler}
\end{equation}
where $c$ denotes the text condition. We measure denoising cost by the number of function evaluations (NFE), defined as the number of denoising steps at which the transformer is evaluated; for models with classifier-free guidance (CFG)~\citep{cfg}, each such step includes both passes.


\vspace{-5pt}

\subsection{Forecasting for Diffusion Acceleration}
\label{sec:bg_forecasting}
\vspace{-5pt}

Adjacent denoising steps produce highly correlated features and outputs, allowing redundant computation to be avoided \citep{deepcache,fora,taylorseer,spectrum}. Forecasting exploits this temporal redundancy by predicting skipped signals from past observations. Let $\mathcal{U}\subseteq\{0,\ldots,N-1\}$ denote the steps that evaluate the denoiser. For output-level forecasting,
\begin{equation}
\widehat{Y}_j=f(h_j,j),\qquad j\notin\mathcal{U},
\label{eq:forecast}
\end{equation}
where $h_j$ encodes past outputs and $\widehat{Y}_j$ replaces $Y_j$ in the sampler. Forecasting may instead target intermediate features, making the choice of forecast target and representation a key design decision.


\vspace{-5pt}

\subsection{Kalman Filter}
\label{sec:bg_kalman}
\vspace{-5pt}

The Kalman filter \citep{kalman1960} is a classical method that recursively estimates the state of a dynamic signal from intermittent observations. It models the state transition as
\begin{equation}
h_j=A_jh_{j-1}+w_j,
\qquad
w_j\sim\mathcal{N}(0,Q_j),
\label{eq:kf_dynamics}
\end{equation}
\begin{equation}
y_j=H_jh_j+\eta_j,
\qquad
\eta_j\sim\mathcal{N}(0,R_j),
\label{eq:kf_observation}
\end{equation}
where $A_j$ is the state-transition matrix, $H_j$ maps the state to the observation, and $Q_j$ and $R_j$ are the process- and observation-noise covariances, respectively. The filter alternates between \emph{prediction} and \emph{update}, with updates only when observations are available. Given the previous estimate $\widehat{h}_{j-1}$ and covariance $P_{j-1}$, it predicts
\begin{equation}
\overline{h}_j=A_j\widehat{h}_{j-1},
\qquad
\overline{P}_j=A_jP_{j-1}A_j^\top+Q_j.
\label{eq:kf_predict}
\end{equation}
When an observation $y_j$ is available, the prediction is corrected using
\begin{equation}
K_j
=
\overline{P}_jH_j^\top
\left(H_j\overline{P}_jH_j^\top+R_j\right)^{-1},
\end{equation}
\begin{equation}
\widehat{h}_j
=
\overline{h}_j+K_j(y_j-H_j\overline{h}_j),
\qquad
P_j=(I-K_jH_j)\overline{P}_j.
\label{eq:kf_update}
\end{equation}
Here, $K_j$ controls the balance between the new observation and the model prediction. An important property of the Kalman filter is that it is fully online, requiring a constant number of operations and no window of past observations to store or refit, so a large number of signals can be tracked in parallel at negligible cost. Between observations, the filter advances the state, while the covariance grows to reflect increasing forecast uncertainty, making the filter suitable for forecasting over extended gaps between observations.



\vspace{-5pt}

\section{Method}
\label{sec:method}

\vspace{-5pt}


\textit{\textbf{Spec}tral \textbf{K}alman Fore\textbf{Cast}ing} (SpecKCast) is an inference-time acceleration method for DiTs that reduces transformer evaluations during denoising. Unlike feature-caching methods that approximate an intermediate activation and still execute the remaining network layers, SpecKCast models the final velocity consumed by the sampler. SpecKCast applies a 2D DCT to the model output and forecasts each frequency coefficient individually, while partitioning coefficients into bands that share forecast uncertainty and a noise-schedule-derived time coordinate, warped nonlinearly to better align with each band's temporal evolution. SpecKCast uses a Kalman filter to track each DCT coefficient's value and temporal derivative, while sharing covariance statistics within each frequency band. The predicted coefficients are transformed back to the spatial domain and passed directly to the sampler.


\vspace{-5pt}
\subsection{Frequency-Aware Forecasting}
\vspace{-3pt}

\subsubsection{Spatial-frequency Decomposition}
\label{sec:freq_decomp}

Forecasting the entire transformer output with a single temporal model ignores an important property of diffusion predictions: different spatial frequencies exhibit different temporal dynamics, as shown in \Cref{fig:fig1_velocity}. Low-frequency content settles early and drifts slowly, while high-frequency content remains active until the end of sampling. To account for these distinct dynamics, SpecKCast transforms each output into the spatial-frequency domain and forecasts its coefficients separately.

Given the denoiser output $Y_j \in \mathbb{R}^{C \times H \times W}$ at timestep $j$, we apply an orthonormal 2D DCT independently to each channel, $Z_j = \operatorname{DCT}_{2\mathrm{D}}(Y_j)$, and partition the frequency plane into $L$ non-overlapping radial bands. Each frequency coefficient is treated as a separate temporal signal and forecast by its own predictor, while the band determines how each coefficient's dynamics are modeled. On steps where the transformer computes the full denoising step, the observed $Z_j$ updates the predictors; on skipped steps, the forecasters extrapolate each coefficient's trajectory and together form $\widehat{Z}_j$. The forecast is then transformed back to the spatial domain, $\widehat{Y}_j = \operatorname{IDCT}_{2\mathrm{D}}(\widehat{Z}_j)$, and passed directly to the diffusion sampler. 


\vspace{-5pt}
\subsubsection{Band-specific Denoising Time}
\label{sec:warp}

Each coefficient in SpecKCast is forecast as a function of time, which raises the question of what to use as the time variable. A natural choice is the step index $j$, but it does not capture the nonuniform progression of the denoising process. As a result, equal increments of $j$ can correspond to very different amounts of change, making the trajectory harder to forecast. We instead derive the time variable from the \emph{noise schedule}, expressed as the log signal-to-noise ratio $\lambda_j = 2\log\!\big((1-\sigma_j)/\sigma_j\big)$. However, individual frequency bands still change most during different phases of denoising. A band's trajectory therefore bends where the band is most active and stays flat where it is not, making it nonlinear in $\lambda_j$. What the forecaster needs is a time variable whose increments reflect how much each band is expected to evolve. We therefore warp $\lambda_j$ separately for each band $\ell$,
\begin{equation}
t_\ell(j) = \operatorname{sigmoid}\!\Big(\frac{\lambda_j + \log p_\ell}{T_w}\Big),
\qquad
s_\ell(j) = \operatorname{rescale}_{[-1,1]}\big(t_\ell(j)\big),
\label{eq:warp}
\end{equation}
where $p_\ell$ is the calibrated signal power of band $\ell$ (\Cref{sec:calib}) and $T_w$ sets the width of each band's active window. The rescaling maps the first and last values of $t_\ell$ to $-1$ and $1$, so all bands share the same range. The resulting coordinate advances more when the band is expected to evolve and less when it is stable. This transformation only changes the forecasting coordinate and leaves the diffusion process unchanged.



\vspace{-5pt}
\subsection{Spectral Kalman Forecasting}
\label{sec:kalman}
\vspace{-3pt}

SpecKCast uses a Kalman filter to track each frequency coefficient with a stepwise discrete dynamics model. The state of coefficient $i$ at step $j$ is $h_{i,j}=(m_{i,j},\nu_{i,j})$, where $m_{i,j}$ is its estimated value and $\nu_{i,j}$ is its temporal derivative. Each coefficient maintains its own state, while the time coordinate $s_\ell$, covariance, and process-noise statistics are shared within each band. Over each adjacent time increment $\Delta_{\ell,j}=s_\ell(j)-s_\ell(j-1)$, the state evolves
{
\setlength{\abovedisplayskip}{2pt}
\setlength{\belowdisplayskip}{6pt}
\setlength{\abovedisplayshortskip}{2pt}
\setlength{\belowdisplayshortskip}{6pt}
\begin{equation}
m_{i,j}
=
m_{i,j-1}+\Delta_{\ell,j}\nu_{i,j-1},
\qquad
\nu_{i,j}
=
\rho_{\ell,j}\nu_{i,j-1},
\qquad
\rho_{\ell,j}
=
e^{-\Delta_{\ell,j}/\tau_\ell},
\label{eq:dlt}
\end{equation}
}
where $\tau_\ell$ controls how quickly the temporal derivative decays. This prediction is applied at every sampler step, including skipped ones, while the Kalman update is applied only at computed steps. The value $m_{i,j}$ advances according to the temporal derivative $\nu_{i,j}$, which gradually decays. This damping reduces the influence of outdated temporal changes and limits overshooting over extended skips. To incorporate these dynamics into the Kalman filter, we define the transition matrix $A_{\ell,j}$ and process-noise covariance $Q_{\ell,j}$ as
{
\setlength{\abovedisplayskip}{5pt}
\setlength{\belowdisplayskip}{8pt}
\setlength{\abovedisplayshortskip}{5pt}
\setlength{\belowdisplayshortskip}{8pt}
\begin{equation}
A_{\ell,j}
=
\begin{bmatrix}
1 & \Delta_{\ell,j}\\
0 & \rho_{\ell,j}
\end{bmatrix},
\qquad
Q_{\ell,j}
=
q_{\ell,j}
\begin{bmatrix}
\Delta_{\ell,j}^3/3 & \Delta_{\ell,j}^2/2\\
\Delta_{\ell,j}^2/2 & \Delta_{\ell,j}
\end{bmatrix},
\label{eq:spectral_kalman_matrices}
\end{equation}
}
Here, $q_{\ell,j}$ is the process-noise intensity, calibrated from the temporal activity of band $\ell$ and adapted during inference using observed forecast errors. The construction of $A_{\ell,j}$ and $Q_{\ell,j}$ is detailed in \Cref{app:kalman_matrices}. The covariance shared by coefficients in band $\ell$ evolves according to
{
\setlength{\abovedisplayskip}{4pt}
\setlength{\belowdisplayskip}{6pt}
\setlength{\abovedisplayshortskip}{4pt}
\setlength{\belowdisplayshortskip}{6pt}
\begin{equation}
P_{\ell,j}
=
A_{\ell,j}P_{\ell,j-1}A_{\ell,j}^{\top}
+
Q_{\ell,j}.
\label{eq:spectral_covariance}
\end{equation}
}
Initially, we set the state as $m_{i,0}=Z_{i,0}$ and $\nu_{i,0}=0$. On compute steps, the observed coefficient $Z_{i,j}$ corrects the predicted state through the Kalman update. On skipped steps, the predicted coefficient values $m_{i,j}$ form the spectral forecast $\widehat{Z}_j$, which is transformed back to the spatial domain and passed to the sampler. Additionally, a separate covariance tracker estimates how each band's uncertainty grows across skipped steps. A compute step is triggered when the combined uncertainty exceeds a threshold.


\subsection{Uncertainty-Aware Scheduling}
\label{sec:scheduling}

The remaining question is which steps to compute. A fixed schedule is the simplest choice, but it cannot respond to changes in forecasting difficulty across generations and denoising phases. SpecKCast instead schedules adaptively, computing a step when the uncertainty accumulated by the Kalman filter indicates that the forecast is becoming unreliable. This requires combining the per-band covariances into a single decision quantity. Coefficients within the same band share the covariance $P_{\ell,j}$ because they use the same time coordinate and process-noise model. This avoids maintaining a separate covariance for every coefficient and reduces uncertainty tracking to only $L$ matrices. We therefore aggregate the per-band uncertainties into a scalar \emph{risk},
{
\setlength{\abovedisplayskip}{4pt}
\setlength{\belowdisplayskip}{6pt}
\setlength{\abovedisplayshortskip}{4pt}
\setlength{\belowdisplayshortskip}{6pt}
\begin{equation}
R(j)
=
\omega(j)
\sum_{\ell=1}^{L}
d_\ell P_{11}^{(\ell)}(j),
\qquad
\omega(j)
=
\exp\left(-\beta\frac{j}{N-1}\right),
\label{eq:risk}
\end{equation}
}
where $d_\ell$ is the number of coefficients in band $\ell$ and $\beta$ controls how strongly late-stage risk is attenuated. Here, $P^{(\ell)}_{11}(j)$ is the top-left entry of $P_{\ell,j}$, tracked by the separate recursion described in \cref{app:sched_sidecar}. Since the rows and columns of $P_{\ell,j}$ follow the state ordering $h_{i,j}=(m_{i,j},\nu_{i,j})$, this entry is the predicted variance of the error in the coefficient estimate $m_{i,j}$. Multiplying it by $d_\ell$ approximates the total forecast uncertainty of band $\ell$. Accordingly, scheduling follows a simple threshold rule. After each compute step, we propagate the covariance forward to predict the risk at future steps. SpecKCast uses forecasted outputs until the first step $k$ satisfying $R(k)>\varepsilon$, at which point the transformer is evaluated. A smaller $\varepsilon$ triggers more frequent evaluations, while a larger value permits longer skips. We calibrate $\varepsilon$ for the desired compute budget, while the covariance adapts to the observed forecast errors during generation.




\vspace{-5pt}
\section{Experiments}
\label{sec:experiment}
\vspace{-3pt}

\paragraph{Setup.}
We evaluate SpecKCast on two state-of-the-art text-to-image models,
FLUX.1-dev~\citep{flux} at $1024\times1024$ and Qwen-Image~\citep{qwen} at
$1328\times1328$, and further on two text-to-video models,
Wan2.1-14B~\citep{wan} and HunyuanVideo~\citep{hunyuanvideo}. Following common
practice, we use 50-step sampling as the full-computation reference for all
models. We compare against FORA~\citep{fora}, TeaCache~\citep{teacache},
HiCache~\citep{hicache}, SeaCache~\citep{seacache}, TaylorSeer~\citep{taylorseer},
and Spectrum~\citep{spectrum}. Among them, TaylorSeer and Spectrum are the most
closely related to our method, as they also forecast future features across
timesteps instead of directly reusing cached ones. All baselines are evaluated
using their official implementations. We decompose each prediction into $L=7$ frequency bands and, unless otherwise stated, set $\tau=0.1$, $T_w=2.0$, and $\beta=0.6$. For models that use classifier-free guidance~\citep{cfg}, SpecKCast forecasts the guided velocity (\cref{app:cfg}).
All experiments are conducted on NVIDIA H100 GPUs.

\begin{table}[t]
  \centering
  \caption{Quantitative comparison in text-to-image generation on FLUX.1-dev and Qwen-Image.}
  \label{tab:image-maintable}
  \vspace{0.4em}
  \setlength{\tabcolsep}{3pt}
  \renewcommand{\arraystretch}{1.05}
  \resizebox{\textwidth}{!}{%
    \begin{tabular}{clcccccccc!{\color{black}\vrule}cccccccc}
    \toprule
  & & \multicolumn{8}{c!{\color{black}\vrule}}{\textbf{FLUX.1-dev} (1024$\times$1024)}
  & \multicolumn{8}{c}{\textbf{Qwen-Image} (1328$\times$1328)} \\
\cmidrule(lr){3-10}\cmidrule(lr){11-18}

  & & \multicolumn{2}{c}{Acceleration}
  & \hdrspan{Mem.\ $\Delta$\\{\small (GiB)}$\downarrow$}
  & \multicolumn{3}{c}{Quality}
  & \hdrspan{Image\\Reward$\uparrow$}
  & \hdrspan{CLIP$\uparrow$}
  & \multicolumn{2}{c}{Acceleration}
  & \hdrspan{Mem.\ $\Delta$\\{\small (GiB)}$\downarrow$}
  & \multicolumn{3}{c}{Quality}
  & \hdrspan{Image\\Reward$\uparrow$}
  & \hdrspan{CLIP$\uparrow$} \\
\cmidrule(lr){3-4}\cmidrule(lr){6-8}
\cmidrule(lr){11-12}\cmidrule(lr){14-16}

  \textbf{NFE} & \textbf{Method}
  & Latency(s)$\downarrow$ & Speedup$\uparrow$ &
  & PSNR$\uparrow$ & SSIM$\uparrow$ & LPIPS$\downarrow$ & &
  & Latency(s)$\downarrow$ & Speedup$\uparrow$ &
  & PSNR$\uparrow$ & SSIM$\uparrow$ & LPIPS$\downarrow$ & & \\
\midrule

    $50$
      & Reference
      & 11.68 & 1.00 & -- & -- & -- & -- & 0.98 & 27.20
      & 52.22 & 1.00 & -- & -- & -- & -- & 1.24 & 29.14 \\
    \midrule

    \multirow{8}{*}{$\approx14$}
      & Base
      & 3.35 & 3.48 & 0.00 & 15.30 & 0.633 & 0.423 & 0.98 & 27.51
      & 14.83 & 3.52 & 0.00 & 15.18 & 0.640 & 0.393 & 1.08 & 29.19 \\
      & FORA
      & 3.98 & 2.93 & 2.01 & 15.03 & 0.628 & 0.433 & 0.96 & 27.48
      & 16.11 & 3.24 & 9.46 & 14.03 & 0.571 & 0.456 & 1.02 & 28.97 \\
      & TeaCache
      & 3.55 & 3.29 & 0.07 & 16.86 & 0.684 & 0.355 & \textbf{1.01} & \textbf{27.57}
      & 15.68 & 3.33 & 0.20 & 19.42 & 0.781 & 0.229 & 1.19 & \textbf{29.36} \\
      & HiCache
      & 4.42 & 2.64 & 4.02 & 18.02 & 0.708 & 0.292 & 1.00 & 27.50
      & 19.16 & 2.72 & 18.97 & 16.04 & 0.647 & 0.335 & 1.16 & 29.08 \\
      & SeaCache
      & 3.59 & 3.25 & 0.09 & 22.30 & 0.813 & 0.184 & 0.98 & 27.40
      & 15.78 & 3.31 & 0.24 & 21.43 & 0.821 & 0.190 & 1.20 & 29.11 \\
      & TaylorSeer
      & 4.59 & 2.55 & 4.02 & 17.74 & 0.700 & 0.305 & 1.00 & 27.32
      & 18.37 & 2.84 & 18.97 & 14.22 & 0.598 & 0.423 & 1.05 & 28.58 \\
      & Spectrum
      & 3.49 & 3.36 & 1.15 & 23.60 & 0.814 & 0.165 & 0.98 & 27.33
      & 15.72 & 3.32 & 2.84 & 23.66 & 0.851 & 0.135 & 1.21 & 29.16 \\
    \rowcolor{gray!15}
      & \textit{SpecKCast}
      & \textbf{3.45} & \textbf{3.38} & \textbf{0.005}
      & \textbf{24.28} & \textbf{0.836} & \textbf{0.146} & 0.98 & 27.32
      & \textbf{15.01} & \textbf{3.48} & \textbf{0.01}
      & \textbf{24.16} & \textbf{0.852} & \textbf{0.134}
      & \textbf{1.22} & 29.17 \\
    \midrule

    \multirow{8}{*}{$\approx10$}
      & Base
      & 2.48 & 4.71 & 0.00 & 14.55 & 0.599 & 0.483 & 0.87 & 27.31
      & 10.60 & 4.93 & 0.00 & 14.03 & 0.586 & 0.480 & 0.95 & 28.89 \\
      & FORA
      & 3.16 & 3.70 & 2.01 & 14.32 & 0.591 & 0.502 & 0.84 & 27.40
      & 12.16 & 4.29 & 9.46 & 12.73 & 0.502 & 0.561 & 0.65 & 28.15 \\
      & TeaCache
      & 2.56 & 4.57 & 0.07 & 15.66 & 0.631 & 0.440 & 0.97 & 27.50
      & 11.46 & 4.56 & 0.20 & 16.11 & 0.656 & 0.390 & 1.06 & 28.84 \\
      & HiCache
      & 3.62 & 3.23 & 4.02 & 16.30 & 0.635 & 0.390 & 0.99 & 27.60
      & 15.41 & 3.39 & 18.97 & 12.77 & 0.488 & 0.519 & 0.88 & 28.02 \\
      & SeaCache
      & 2.55 & 4.59 & 0.09 & 19.29 & 0.731 & 0.304 & 0.97 & 27.52
      & 11.75 & 4.45 & 0.24 & 19.61 & 0.767 & 0.257 & 1.15 & 29.08 \\
      & TaylorSeer
      & 3.55 & 3.29 & 4.02 & 15.28 & 0.607 & 0.436 & 0.97 & 27.37
      & 14.40 & 3.63 & 18.97 & 10.79 & 0.405 & 0.623 & 0.62 & 26.85 \\
      & Spectrum
      & 2.55 & 4.59 & 0.87 & 21.27 & 0.729 & 0.261 & 0.98 & 27.62
      & 11.53 & 4.53 & 2.21 & 20.85 & 0.746 & 0.318 & 1.17 & \textbf{29.80} \\
    \rowcolor{gray!15}
      & \textit{SpecKCast}
      & \textbf{2.46} & \textbf{4.74} & \textbf{0.005}
      & \textbf{21.77} & \textbf{0.749} & \textbf{0.258} & 0.98 & 27.57
      & \textbf{10.80} & \textbf{4.83} & \textbf{0.01}
      & \textbf{22.28} & \textbf{0.790} & \textbf{0.201}
      & \textbf{1.19} & 29.01 \\
    \midrule

    \multirow{8}{*}{$\approx8$}
      & Base
      & 1.98 & 5.89 & 0.00 & 14.29 & 0.578 & 0.518 & 0.81 & 27.39
      & 8.54 & 6.11 & 0.00 & 13.53 & 0.561 & 0.531 & 0.80 & 28.78 \\
      & FORA
      & 2.70 & 4.33 & 2.01 & 14.12 & 0.574 & 0.529 & 0.74 & 27.07
      & 10.15 & 5.15 & 9.46 & 12.39 & 0.475 & 0.589 & 0.40 & 27.76 \\
      & TeaCache
      & 2.13 & 5.75 & 0.07 & 15.07 & 0.600 & 0.506 & 0.82 & 27.48
      & 11.50 & 4.54 & 0.20 & 16.00 & 0.648 & 0.397 & 1.04 & 28.83 \\
      & HiCache
      & 3.23 & 3.48 & 4.02 & 15.13 & 0.582 & 0.466 & \textbf{0.98} & 27.53
      & 13.50 & 3.87 & 18.97 & 10.86 & 0.391 & 0.639 & 0.56 & 27.21 \\
      & SeaCache
      & 2.13 & 5.48 & 0.09 & 18.47 & \textbf{0.688} & 0.370 & 0.96 & 27.52
      & 10.38 & 5.03 & 0.24 & 18.41 & 0.733 & 0.307 & 1.09 & 28.95 \\
      & TaylorSeer
      & 3.14 & 3.72 & 4.02 & 13.80 & 0.543 & 0.516 & 0.93 & 27.16
      & 12.48 & 4.18 & 18.97 & 9.17 & 0.322 & 0.720 & 0.26 & 25.90 \\
      & Spectrum
      & 2.15 & 5.43 & 0.73 & 18.71 & 0.632 & 0.397 & 0.90 & 27.73
      & 10.23 & 5.48 & 1.89 & 16.91 & 0.604 & 0.527 & 0.90 & \textbf{29.68} \\
    \rowcolor{gray!15}
      & \textit{SpecKCast}
      & \textbf{2.11} & \textbf{5.54} & \textbf{0.005}
      & \textbf{19.81} & 0.683 & \textbf{0.353} & 0.89 & 27.48
      & \textbf{8.70} & \textbf{6.00} & \textbf{0.01}
      & \textbf{20.44} & \textbf{0.740} & \textbf{0.267}
      & \textbf{1.11} & 28.93 \\
    \bottomrule
    \end{tabular}%
  }
\end{table}

\begin{figure}[t]
  \centering
  \includegraphics[width=\textwidth]{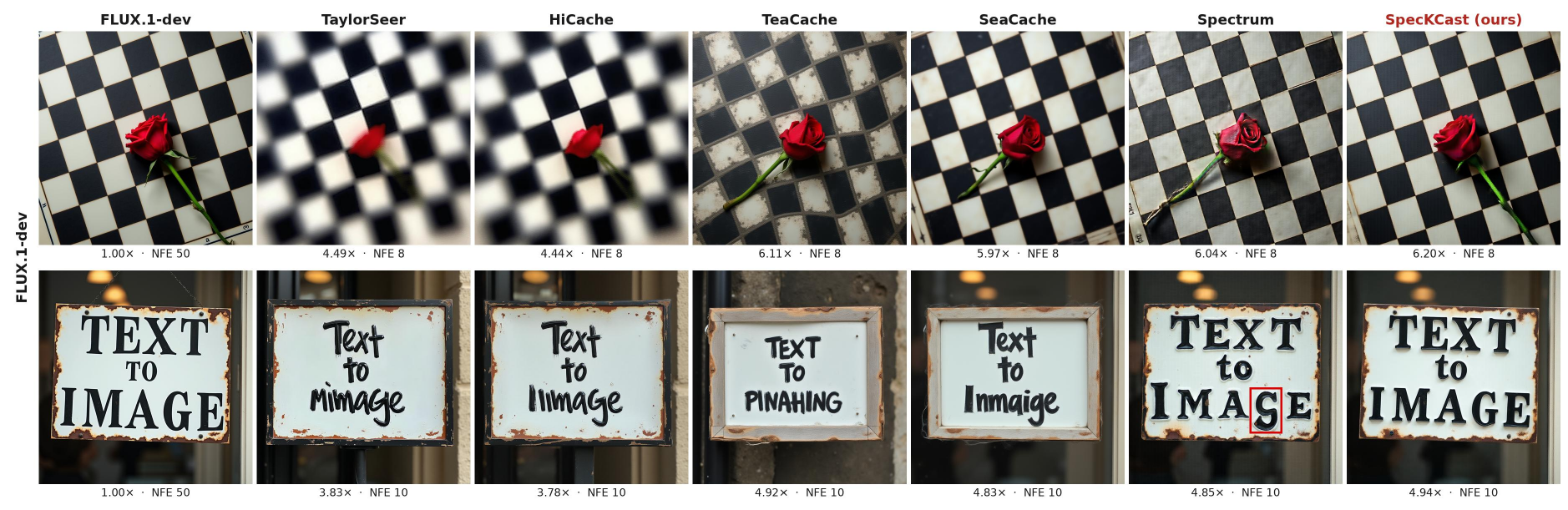}
  \vspace{-15pt}  
  \caption{Qualitative comparison on text-to-image generation using FLUX.1-dev.}
  \label{fig:flux-qualitative}
  \vspace{-10pt}  
\end{figure}


\begin{table*}[t]
\centering
\caption{Quantitative comparison in text-to-video generation on Wan2.1-14B and HunyuanVideo.}
\label{tab:video_main}
\setlength{\tabcolsep}{4pt}
\renewcommand{\arraystretch}{1.05}
\resizebox{\textwidth}{!}{%
\begin{tabular}{clccccccc!{\color{black}\vrule}ccccccc}
\toprule
& & \multicolumn{7}{c!{\color{black}\vrule}}{\textbf{Wan2.1-14B}}
& \multicolumn{7}{c}{\textbf{HunyuanVideo}} \\
\cmidrule(lr){3-9}\cmidrule(lr){10-16}

& & \multicolumn{2}{c}{Acceleration}
& \hdrspan{Mem.\ $\Delta$\\{\small (GiB)}$\downarrow$}
& \multicolumn{3}{c}{Quality}
& \hdrspan{VBench\\Quality$\uparrow$}
& \multicolumn{2}{c}{Acceleration}
& \hdrspan{Mem.\ $\Delta$\\{\small (GiB)}$\downarrow$}
& \multicolumn{3}{c}{Quality}
& \hdrspan{VBench\\Quality$\uparrow$} \\
\cmidrule(lr){3-4}\cmidrule(lr){6-8}
\cmidrule(lr){10-11}\cmidrule(lr){13-15}

\textbf{NFE} & \textbf{Method}
& Latency(s)$\downarrow$ & Speedup$\uparrow$ &
& PSNR$\uparrow$ & SSIM$\uparrow$ & LPIPS$\downarrow$ &
& Latency(s)$\downarrow$ & Speedup$\uparrow$ &
& PSNR$\uparrow$ & SSIM$\uparrow$ & LPIPS$\downarrow$ & \\
\midrule

50 & Reference
& 370 & 1.00 & -- & -- & -- & -- & 81.11
& 183 & 1.00 & -- & -- & -- & -- & 80.26 \\
\midrule
 
\multirow{9}{*}{$\approx 14$}
 & Base
& 116 & 3.20 & 0.00 & 13.95 & 0.471 & 0.518 & 79.81
& 58.3 & 3.14 & 0.00 & 17.33 & 0.641 & 0.417 & 78.38 \\
 & FORA
& 108.3 & 3.42 & 0.62 & 21.62 & 0.753 & 0.192 & 80.63
& 56.8 & 3.22 & 0.19 & 23.46 & 0.775 & 0.216 & 78.90 \\
 & TeaCache
& \textbf{105.2} & \textbf{3.52} & 0.94 & 22.13 & 0.772 & 0.191 & 80.24
& 61.2 & 2.99 & 0.37 & 20.56 & 0.709 & 0.279 & 79.35 \\
 & HiCache
& 111.9 & 3.31 & 3.75 & 18.42 & 0.675 & 0.267 & 80.12
& 56.8 & 3.22 & 1.12 & 23.43 & 0.785 & 0.198 & 79.54 \\
 & SeaCache
& 113.2 & 3.27 & 4.37 & 23.10 & 0.795 & 0.157 & 80.38
& 71.3 & 2.57 & 0.37 & 27.59 & 0.864 & 0.121 & 79.53 \\
 & TaylorSeer
& 109.4 & 3.38 & 4.69 & 17.94 & 0.667 & 0.264 & 80.12
& 57.2 & 3.20 & 1.49 & 22.29 & 0.767 & 0.217 & \textbf{79.98} \\
 & Spectrum
& 110.3 & 3.35 & 13.12 & 22.93 & 0.789 & \textbf{0.136} & 80.76
& 58.0 & 3.15 & 3.56 & 26.95 & 0.831 & 0.133 & 79.32 \\
\rowcolor{gray!15}
 & \textit{SpecKCast}
& \textbf{105.2} & \textbf{3.52} & \textbf{0.05} & \textbf{24.47} & \textbf{0.813} & 0.140 & \textbf{80.86}
& \textbf{55.0} & \textbf{3.33} & \textbf{0.03} & \textbf{29.72} & \textbf{0.884} & \textbf{0.107} & 79.68 \\
\midrule
 
\multirow{8}{*}{$\approx 10$}
 & Base
& 80.50 & 4.59 & 0.00 & 13.66 & 0.452 & 0.578 & 78.05
& 42.7 & 4.28 & 0.00 & 16.73 & 0.621 & 0.469 & 76.59 \\
 & FORA
& 79.6 & 4.65 & 0.62 & 20.68 & 0.709 & 0.270 & 79.76
& 43.0 & 4.26 & 0.19 & 22.03 & 0.726 & 0.306 & 77.12 \\
 & TeaCache
& 80.3 & 4.61 & 0.94 & 21.11 & 0.732 & 0.245 & 79.05
& 47.2 & 3.88 & 0.37 & 19.65 & 0.672 & 0.329 & 78.19 \\
 & HiCache
& 80.3 & 4.61 & 3.75 & 14.04 & 0.519 & 0.474 & 74.96
& 43.0 & 4.26 & 1.12 & 21.17 & 0.718 & 0.293 & 78.06 \\
 & SeaCache
& 89.5 & 4.13 & 4.37 & 19.53 & 0.686 & 0.271 & 79.87
& 45.4 & 4.03 & 0.37 & 19.34 & 0.698 & 0.338 & 78.08 \\
 & TaylorSeer
& 80.2 & 4.61 & 4.69 & 14.57 & 0.549 & 0.404 & 76.20
& 43.4 & 4.22 & 1.49 & 19.54 & 0.689 & 0.318 & 78.19 \\
 & Spectrum
& 80.8 & 4.58 & 9.38 & 21.41 & 0.733 & \textbf{0.175} & 80.42
& 43.8 & 4.17 & 2.81 & 24.60 & 0.767 & 0.191 & 78.80 \\
\rowcolor{gray!15}
 & \textit{SpecKCast}
& \textbf{79.1} & \textbf{4.68} & \textbf{0.05} & \textbf{21.62} & \textbf{0.750} & 0.181 & \textbf{80.48}
& \textbf{42.7} & \textbf{4.29} & \textbf{0.03} & \textbf{25.62} & \textbf{0.821} & \textbf{0.172} & \textbf{79.25} \\
\midrule
 
\multirow{9}{*}{$\approx 8$}
 & Base
& 80.2 & 4.61 & 0.00 & 13.56 & 0.449 & 0.614 & 76.98
& 38.2 & 4.79 & 0.00 & 16.35 & 0.612 & 0.510 & 75.01 \\
 & FORA
& 65.5 & 5.65 & 0.62 & \textbf{20.19} & 0.669 & 0.336 & 78.38
& 36.2 & 5.06 & 0.19 & 21.22 & 0.655 & 0.346 & 75.00 \\
 & TeaCache
& 65.8 & 5.62 & 0.94 & 19.61 & 0.656 & 0.335 & 76.52
& 36.9 & 4.96 & 0.37 & 18.37 & 0.617 & 0.415 & 76.00 \\
 & HiCache
& 66.3 & 5.58 & 3.75 & 12.06 & 0.430 & 0.578 & 72.01
& 36.2 & 5.06 & 1.12 & 19.83 & 0.643 & 0.346 & 75.42 \\
 & SeaCache
& \textbf{65.0} & \textbf{5.69} & 4.37 & 15.27 & 0.519 & 0.482 & 76.97
& 37.0 & 4.95 & 0.37 & 18.98 & 0.681 & 0.380 & \textbf{76.81} \\
 & TaylorSeer
& 65.7 & 5.63 & 4.69 & 13.02 & 0.480 & 0.475 & 73.57
& 36.5 & 5.01 & 1.49 & 18.15 & 0.620 & 0.387 & 74.90 \\
 & Spectrum
& 66.0 & 5.61 & 7.50 & 17.81 & 0.638 & 0.303 & 78.74
& 37.9 & 4.82 & 2.43 & 21.74 & 0.674 & 0.297 & 75.77 \\
\rowcolor{gray!15}
 & \textit{SpecKCast}
& 65.3 & 5.67 & \textbf{0.05} & 19.94 & \textbf{0.683} & \textbf{0.271} & \textbf{79.20}
& \textbf{35.8} & \textbf{5.12} & \textbf{0.03} & \textbf{22.34} & \textbf{0.747} & \textbf{0.295} & 76.32 \\
\bottomrule
\end{tabular}%
}
\vspace{0.3em}
\end{table*}

\begin{figure}[t]
  \centering
  \includegraphics[width=\textwidth]{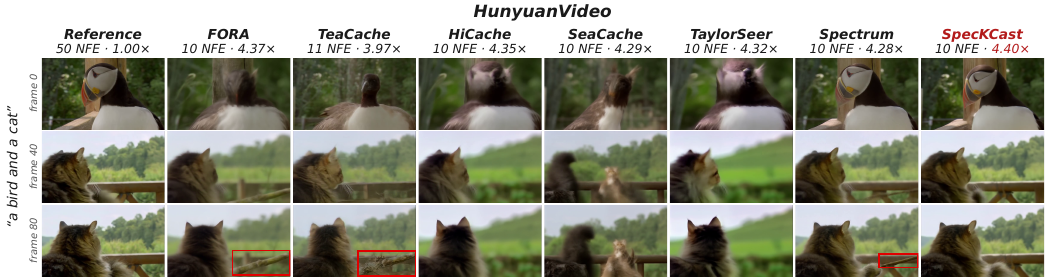}
  \vspace{-15pt} 
  \caption{Qualitative comparison on text-to-video generation using HunyuanVideo.}
  \label{fig:hunyuan-qualitative}
\end{figure}

\paragraph{Evaluation.}
For text-to-image generation, we use the DrawBench~\citep{drawbench} prompts and report ImageReward~\citep{imagereward} and CLIP Score~\citep{clipscore}, which measures text--image alignment. For text-to-video generation, we follow the standard prompts and protocol of VBench~\citep{vbench} and report its Quality Score, which aggregates dimensions such as temporal consistency, motion smoothness, and per-frame aesthetic and imaging quality. Since an accelerated sampler should reproduce the output of the full model rather than merely produce a plausible sample, we also compare each accelerated sample with its 50-step reference generated from the same prompt and seed. We report PSNR for pixel-level error, SSIM~\citep{ssim} for structural similarity, and LPIPS~\citep{lpips} for perceptual distance. For efficiency, we report sampling latency, speedup over the 50-step reference, and memory overhead (Mem.~$\Delta$), defined as the peak allocated GPU memory above that of Base at the same NFE.

\begin{table}[t]
\centering
\scriptsize
\caption{Ablation of SpecKCast components on FLUX.1-dev at 14 NFE.}
\label{tab:ablation-main}
\vspace{0.4em}
\setlength{\tabcolsep}{3pt}
\renewcommand{\arraystretch}{1.05}
\resizebox{\textwidth}{!}{%
\begin{tabular}{@{} lllc | cccccc @{}}
\toprule
Forecaster
& Scheduler
& Time variable
& \shortstack{Frequency\\decomposition}
& Latency(s)$\downarrow$
& PSNR$\uparrow$
& SSIM$\uparrow$
& LPIPS$\downarrow$
& \shortstack{Image\\Reward$\uparrow$}
& CLIP$\uparrow$ \\
\midrule

Taylor & fixed & warped-logSNR
& $\checkmark$ & 3.46 & 21.90 & \underline{0.811}
& \underline{0.171} & \underline{0.978} & 27.24 \\

Chebyshev & fixed & warped-logSNR
& $\checkmark$ & 3.97 & 21.10 & 0.761
& 0.252 & 0.975 & \textbf{27.66} \\

Kalman & fixed  & warped-logSNR
& $\checkmark$ & \textbf{3.43} & \underline{22.09} & 0.810
& 0.178 & 0.973 & 27.27 \\

\midrule

Kalman & adaptive & raw-logSNR
& $\checkmark$ & 3.52 & 21.03 & 0.775
& 0.210 & 0.977 & 27.24 \\

Kalman & adaptive & uniform step index
& $\checkmark$ & 3.51 & 19.54 & 0.740
& 0.256 & 0.969 & 27.28 \\
\midrule

Kalman & adaptive & warped-logSNR
& $\times$ & 3.48 & 21.19 & 0.785
& 0.201 & 0.966 & 27.13 \\

\midrule

Kalman & adaptive & warped-logSNR
& $\checkmark$ & \underline{3.45} & \textbf{24.28} & \textbf{0.836}
& \textbf{0.146} & \textbf{ 0.980} & \underline{27.32} \\

\bottomrule
\end{tabular}%
}
\end{table}

\begin{figure}[t]
  \centering
  \includegraphics[width=\textwidth]{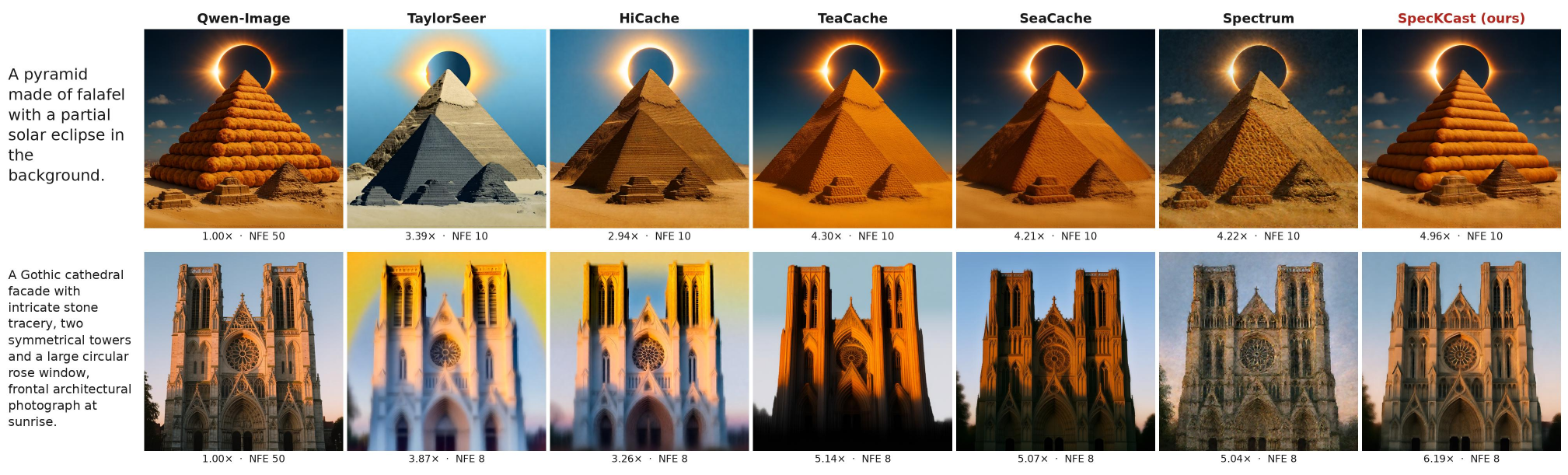}
  \caption{Qualitative comparison on text-to-image generation using Qwen-Image.}
  \label{fig:qwen-qualitative}

  \vspace{2em}

  \includegraphics[width=\textwidth]{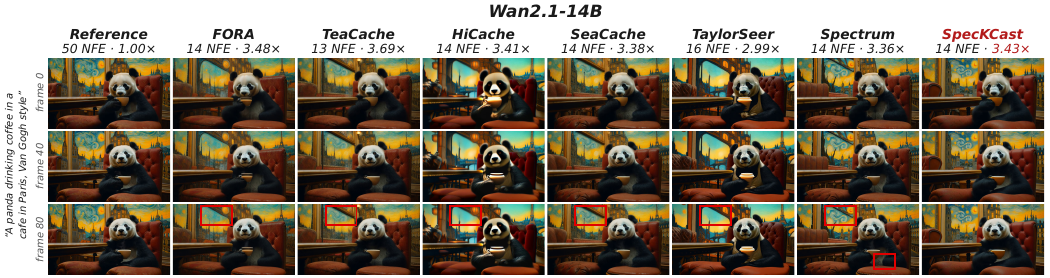}
  \caption{Qualitative comparison on text-to-video generation using Wan2.1-14B.}
  \label{fig:wan-qualitative}
\end{figure}

\subsection{Text-to-image Generation}
\label{sec:t2i}

\Cref{tab:image-maintable} reports results on FLUX.1-dev and Qwen-Image, where Base runs the original sampler with fewer steps to match the NFE. SpecKCast achieves the highest speedup among acceleration methods in all but one setting, along with the best PSNR and LPIPS in all settings and the best SSIM in all but one. At the highest acceleration, it outperforms the strongest baseline by 1.1\,dB PSNR on FLUX.1-dev ($5.54\times$) and 2.0\,dB on Qwen-Image ($6.00\times$). Spectrum, the closest competitor at 14 and 10 NFE, trails SpecKCast by 0.5--1.4\,dB PSNR and degrades sharply on Qwen-Image at 8 NFE, where its LPIPS rises to 0.527 compared with 0.267 for SpecKCast. Compared with Base, which runs at a similar speed, SpecKCast improves PSNR by 5.5--9.0\,dB, showing that it tracks the 50-step trajectory far more closely than simply taking fewer steps. These gains cost at most 0.01\,GiB of memory, whereas TaylorSeer, HiCache, and Spectrum add 0.73--18.97\,GiB. \Cref{fig:flux-qualitative,fig:qwen-qualitative} show qualitative results on FLUX.1-dev and Qwen-Image, respectively. TaylorSeer and HiCache blur the images, TeaCache and SeaCache alter textures and lighting, and every baseline misspells or distorts the text on the sign. SpecKCast stays closest to the 50-step reference while running fastest. More samples are provided in \Cref{app:extended_comparisons}.

\subsection{Text-to-video Generation}
\label{sec:t2v}

\Cref{tab:video_main} reports results on Wan2.1-14B and HunyuanVideo. SpecKCast achieves the highest or near-highest speedup in every setting, along with the best SSIM in all settings and the best PSNR and LPIPS in most. On HunyuanVideo at the 14-NFE budget, it improves PSNR by 2.1\,dB over SeaCache, the strongest baseline, while running at $3.33\times$ instead of $2.57\times$. The advantage holds at the 8-NFE budget, where SpecKCast keeps the best SSIM and LPIPS on both models, while HiCache and TaylorSeer drop to 12.06 and 13.02\,dB PSNR on Wan2.1-14B. SpecKCast also achieves the best VBench Quality on Wan2.1-14B at every budget, staying within 0.25 points of the reference at the 14-NFE budget. Spectrum obtains slightly lower LPIPS on Wan2.1-14B at the 14- and 10-NFE budgets, but requires up to 13.12\,GiB of extra memory, compared with 0.05\,GiB for SpecKCast. \Cref{fig:hunyuan-qualitative,fig:wan-qualitative} show qualitative results on HunyuanVideo and Wan2.1-14B, respectively. On HunyuanVideo, most baselines replace the puffin in the first frame with a blurred or different bird, and SeaCache adds a second cat in later frames. On Wan2.1-14B, the baselines introduce local artifacts (red boxes), mostly in the window background, and HiCache also shifts the lighting. SpecKCast stays closest to the 50-step reference across frames on both models.

\vspace{-5pt}
\subsection{Ablation Studies}
\label{sec:ablation}
\vspace{-5pt}

\Cref{tab:ablation-main} ablates the components of SpecKCast on FLUX.1-dev at 14 NFE. Since uncertainty-aware scheduling relies on the Kalman covariance, we first compare forecasters under a fixed schedule. Kalman and Taylor perform similarly, with Kalman slightly faster and higher in PSNR, while Chebyshev is both slower and less accurate. With our frequency decomposition and time warping, the Taylor forecaster also outperforms TaylorSeer (\Cref{tab:image-maintable}) by 4.2\,dB PSNR while reducing latency from 4.59 to 3.46\,s. Replacing the fixed schedule with uncertainty-aware scheduling (\Cref{sec:scheduling}) improves PSNR by 2.2\,dB. Band-specific time warping (\Cref{sec:warp}) outperforms the raw log-SNR $\lambda_j$ by 3.3\,dB and the step index $j$ by 4.7\,dB, and removing the frequency decomposition lowers PSNR by 3.1\,dB. Overall, the full model achieves the best PSNR, SSIM, LPIPS, and ImageReward. Since all variants except Chebyshev run within 0.1\,s of each other, these gains come at almost no extra latency.

\section{Conclusion}
\label{sec}

We presented SpecKCast, a training-free method for accelerating Diffusion Transformers by forecasting the denoiser output at skipped sampling steps. Rather than modeling the output uniformly, SpecKCast decomposes it into spatial-frequency components and forecasts their distinct temporal dynamics along band-specific, noise-schedule-aligned time coordinates. A lightweight Kalman filter tracks the value and rate of change of each DCT coefficient, while its uncertainty provides a principled signal for adaptively determining when a full transformer evaluation is needed. Across image and video diffusion models, SpecKCast consistently improves the acceleration--quality trade-off over existing caching and forecasting methods, achieving the best fidelity to the 50-step reference in most settings with negligible latency and memory overhead. These results highlight the benefit of explicitly accounting for the distinct temporal evolution of frequency components when forecasting diffusion trajectories. SpecKCast, however, targets only temporal redundancy across denoising steps; each computed step still processes the full token grid and therefore retains the spatial redundancy within a transformer evaluation. Jointly exploiting temporal and spatial redundancy could enable further acceleration and represents a promising direction for future work.



\bibliography{iclr2027_conference}

@article{ho2020denoising,
  title={Denoising diffusion probabilistic models},
  author={Ho, Jonathan and Jain, Ajay and Abbeel, Pieter},
  journal={Advances in neural information processing systems},
  volume={33},
  pages={6840--6851},
  year={2020}
}

@article{liu2022flow,
  title={Flow straight and fast: Learning to generate and transfer data with rectified flow},
  author={Liu, Xingchao and Gong, Chengyue and Liu, Qiang},
  journal={arXiv preprint arXiv:2209.03003},
  year={2022}
}

@article{lipman2022flow,
  title={Flow matching for generative modeling},
  author={Lipman, Yaron and Chen, Ricky TQ and Ben-Hamu, Heli and Nickel, Maximilian and Le, Matt},
  journal={arXiv preprint arXiv:2210.02747},
  year={2022}
}

@article{albergo2025stochastic,
  title={Stochastic interpolants: A unifying framework for flows and diffusions},
  author={Albergo, Michael and Boffi, Nicholas M and Vanden-Eijnden, Eric},
  journal={Journal of Machine Learning Research},
  volume={26},
  number={209},
  pages={1--80},
  year={2025}
}

@article{qwen,
  title={Qwen-image technical report},
  author={Wu, Chenfei and Li, Jiahao and Zhou, Jingren and Lin, Junyang and Gao, Kaiyuan and Yan, Kun and Yin, Sheng-ming and Bai, Shuai and Xu, Xiao and Chen, Yilei and others},
  journal={arXiv preprint arXiv:2508.02324},
  year={2025}
}

@misc{flux,
    author={Black Forest Labs},
    title={FLUX},
    year={2024},
    howpublished={\url{https://github.com/black-forest-labs/flux}},
}

@inproceedings{taylorseer,
  title={From reusing to forecasting: Accelerating diffusion models with taylorseers},
  author={Liu, Jiacheng and Zou, Chang and Lyu, Yuanhuiyi and Chen, Junjie and Zhang, Linfeng},
  booktitle={2025 IEEE/CVF International Conference on Computer Vision (ICCV)},
  pages={15853--15863},
  year={2025},
  organization={IEEE}
}

@article{spectrum,
  title={Adaptive spectral feature forecasting for diffusion sampling acceleration},
  author={Han, Jiaqi and Shi, Juntong and Li, Puheng and Ye, Haotian and Guo, Qiushan and Ermon, Stefano},
  journal={arXiv preprint arXiv:2603.01623},
  year={2026}
}

@article{kalman1960,
  author  = {Kalman, R. E.},
  title   = {A New Approach to Linear Filtering and Prediction Problems},
  journal = {Journal of Basic Engineering},
  volume  = {82},
  number  = {1},
  pages   = {35--45},
  year    = {1960},
  doi     = {10.1115/1.3662552}
}

@inproceedings{deepcache,
  title={Deepcache: Accelerating diffusion models for free},
  author={Ma, Xinyin and Fang, Gongfan and Wang, Xinchao},
  booktitle={2024 IEEE/CVF Conference on Computer Vision and Pattern Recognition (CVPR)},
  pages={15762--15772},
  year={2024},
  organization={IEEE}
}

@inproceedings{peebles2023scalable,
  title={Scalable diffusion models with transformers},
  author={Peebles, William and Xie, Saining},
  booktitle={2023 IEEE/CVF International Conference on Computer Vision (ICCV)},
  pages={4172--4182},
  year={2023},
  organization={IEEE}
}

@article{salimans2022progressive,
  title={Progressive distillation for fast sampling of diffusion models},
  author={Salimans, Tim and Ho, Jonathan},
  journal={arXiv preprint arXiv:2202.00512},
  year={2022}
}

@inproceedings{li2023q,
  title={Q-diffusion: Quantizing diffusion models},
  author={Li, Xiuyu and Liu, Yijiang and Lian, Long and Yang, Huanrui and Dong, Zhen and Kang, Daniel and Zhang, Shanghang and Keutzer, Kurt},
  booktitle={2023 IEEE/CVF International Conference on Computer Vision (ICCV)},
  pages={17489--17499},
  year={2023},
  organization={IEEE}
}

@inproceedings{xia2025training,
  title={Training-free and adaptive sparse attention for efficient long video generation},
  author={Xia, Yifei and Ling, Suhan and Fu, Fangcheng and Wang, Yujie and Li, Huixia and Xiao, Xuefeng and Cui, Bin},
  booktitle={2025 IEEE/CVF International Conference on Computer Vision (ICCV)},
  pages={15982--15993},
  year={2025},
  organization={IEEE}
}

@inproceedings{pixart,
  title={Pixart-$\sigma$: Weak-to-strong training of diffusion transformer for 4k text-to-image generation},
  author={Chen, Junsong and Ge, Chongjian and Xie, Enze and Wu, Yue and Yao, Lewei and Ren, Xiaozhe and Wang, Zhongdao and Luo, Ping and Lu, Huchuan and Li, Zhenguo},
  booktitle={European Conference on Computer Vision},
  pages={74--91},
  year={2024},
  organization={Springer}
}

@inproceedings{cogvideox,
  title={Cogvideox: Text-to-video diffusion models with an expert transformer},
  author={Yang, Zhuoyi and Teng, Jiayan and Zheng, Wendi and Ding, Ming and Huang, Shiyu and Xu, Jiazheng and Yang, Yuanming and Hong, Wenyi and Zhang, Xiaohan and Feng, Guanyu and others},
  booktitle={International Conference on Learning Representations},
  volume={2025},
  pages={83048--83077},
  year={2025}
}

@inproceedings{sdmatte,
  title={Sdmatte: Grafting diffusion models for interactive matting},
  author={Huang, Longfei and Liang, Yu and Zhang, Hao and Chen, Jinwei and Dong, Wei and Chen, Lunde and Liu, Wanyu and Li, Bo and Jiang, Peng-Tao},
  booktitle={2025 IEEE/CVF International Conference on Computer Vision (ICCV)},
  pages={15229--15239},
  year={2025},
  organization={IEEE}
}

@article{sega,
  title={SEGA: Spectral-Energy Guided Attention for Resolution Extrapolation in Diffusion Transformers},
  author={Rajabi, Javad and Shaban, Kimia and Roohi, Koorosh and Lindell, David B and Taati, Babak},
  journal={arXiv preprint arXiv:2605.22668},
  year={2026}
}

@article{proxypose,
  title={ProxyPose: 6-DoF Pose Tracking via Video-to-Video Translation},
  author={Zhang, Ruihang and Taubner, Felix and Ravi, Pooja and Kutulakos, Kiriakos N and Lindell, David B},
  journal={arXiv preprint arXiv:2607.06555},
  year={2026}
}

@inproceedings{meng2023distillation,
  title={On distillation of guided diffusion models},
  author={Meng, Chenlin and Rombach, Robin and Gao, Ruiqi and Kingma, Diederik and Ermon, Stefano and Ho, Jonathan and Salimans, Tim},
  booktitle={2023 IEEE/CVF Conference on Computer Vision and Pattern Recognition (CVPR)},
  pages={14297--14306},
  year={2023},
  organization={IEEE}
}

@article{dmd,
  title={Improved distribution matching distillation for fast image synthesis},
  author={Yin, Tianwei and Gharbi, Micha{\"e}l and Park, Taesung and Zhang, Richard and Shechtman, Eli and Durand, Fredo and Freeman, William T},
  journal={Advances in neural information processing systems},
  volume={37},
  pages={47455--47487},
  year={2024}
}

@inproceedings{yin2024one,
  title={One-step diffusion with distribution matching distillation},
  author={Yin, Tianwei and Gharbi, Micha{\"e}l and Zhang, Richard and Shechtman, Eli and Durand, Fredo and Freeman, William T and Park, Taesung},
  booktitle={2024 IEEE/CVF Conference on Computer Vision and Pattern Recognition (CVPR)},
  pages={6613--6623},
  year={2024},
  organization={IEEE}
}

@inproceedings{song2024improved,
  title={Improved techniques for training consistency models},
  author={Song, Yang and Dhariwal, Prafulla},
  booktitle={International Conference on Learning Representations},
  volume={2024},
  pages={15078--15097},
  year={2024}
}

@article{song2023consistency,
  title={Consistency models},
  author={Song, Yang and Dhariwal, Prafulla and Chen, Mark and Sutskever, Ilya},
  journal={arXiv preprint arXiv:2303.01469},
  year={2023}
}

@inproceedings{meancache,
  title={Meancache: From instantaneous to average velocity for accelerating flow matching inference},
  author={Gao, Huanlin and Chen, Ping and Shi, Fuyuan and Wu, Ruijia and YanTao, Li and Hui, Qiang and You, Yuren and Lu, Ting and Tan, Chao and Zhao, Shaoan and Liu, Zhaoxiang and Zhao, Fang and Wang, Kai and Lian, Shiguo},
  booktitle={International Conference on Learning Representations},
  volume={2026},
  pages={60466--60485},
  year={2026}
}

@inproceedings{hicache,
  title={HiCache: A plug-in scaled-Hermite upgrade for Taylor-style cache-then-forecast diffusion acceleration},
  author={Feng, Liang and Zheng, Shikang and Liu, Jiacheng and Lin, Yuqi and Zhou, Qinming and Cai, Peiliang and Wang, Xinyu and Chen, Junjie and Zou, Chang and Ma, Yue and others},
  booktitle={International Conference on Learning Representations},
  volume={2026},
  pages={112361--112386},
  year={2026}
}

@inproceedings{seacache,
  title={SeaCache: Spectral-Evolution-Aware Cache for Accelerating Diffusion Models},
  author={Chung, Jiwoo and Hyun, Sangeek and Lee, MinKyu and Han, Byeongju and Cha, Geonho and Wee, Dongyoon and Hong, Youngjun and Heo, Jae-Pil},
  booktitle={Proceedings of the IEEE/CVF Conference on Computer Vision and Pattern Recognition},
  pages={14283--14294},
  year={2026}
}

@inproceedings{eigencache,
  title={EigenCache: Rethinking Diffusion Acceleration as Covariance-Optimal Forecasting and Submodular Information Allocation},
  author={Xu, Chenyang and Wang, Dezhen and Chen, Lin and Lin, Kepeng and Wang, Hao},
  booktitle={Forty-third International Conference on Machine Learning},
  year={2026}
}

@article{toca,
  title={Accelerating diffusion transformers with token-wise feature caching},
  author={Zou, Chang and Liu, Xuyang and Liu, Ting and Huang, Siteng and Zhang, Linfeng},
  journal={arXiv preprint arXiv:2410.05317},
  year={2024}
}

@article{fora,
  title={Fora: Fast-forward caching in diffusion transformer acceleration},
  author={Selvaraju, Pratheba and Ding, Tianyu and Chen, Tianyi and Zharkov, Ilya and Liang, Luming},
  journal={arXiv preprint arXiv:2407.01425},
  year={2024}
}

@inproceedings{ertacache,
  title={Ertacache: Error rectification and timesteps adjustment for efficient diffusion},
  author={Peng, Xurui and Yan, Chenqian and Liu, Hong and Ma, Rui and Chen, Fangmin and Wang, Xing and Wu, Zhihua and Liu, Songwei and Lin, Mingbao},
  booktitle={International Conference on Learning Representations},
  volume={2026},
  pages={82104--82120},
  year={2026}
}

@article{delta,
  title={{$\Delta$-DiT}: A Training-Free Acceleration Method Tailored for Diffusion Transformers},
  author={Chen, Pengtao and Shen, Mingzhu and Ye, Peng and Cao, Jianjian and Tu, Chongjun and Bouganis, Christos-Savvas and Zhao, Yiren and Chen, Tao},
  journal={arXiv preprint arXiv:2406.01125},
  year={2024}
}

@inproceedings{teacache,
  title={Timestep embedding tells: It’s time to cache for video diffusion model},
  author={Liu, Feng and Zhang, Shiwei and Wang, Xiaofeng and Wei, Yujie and Qiu, Haonan and Zhao, Yuzhong and Zhang, Yingya and Ye, Qixiang and Wan, Fang},
  booktitle={2025 IEEE/CVF Conference on Computer Vision and Pattern Recognition (CVPR)},
  pages={7353--7363},
  year={2025},
  organization={IEEE}
}

@article{wan,
      title={Wan: Open and Advanced Large-Scale Video Generative Models}, 
      author={Team Wan and Ang Wang and Baole Ai and Bin Wen and Chaojie Mao and Chen-Wei Xie and Di Chen and Feiwu Yu and Haiming Zhao and Jianxiao Yang and Jianyuan Zeng and Jiayu Wang and Jingfeng Zhang and Jingren Zhou and Jinkai Wang and Jixuan Chen and Kai Zhu and Kang Zhao and Keyu Yan and Lianghua Huang and Mengyang Feng and Ningyi Zhang and Pandeng Li and Pingyu Wu and Ruihang Chu and Ruili Feng and Shiwei Zhang and Siyang Sun and Tao Fang and Tianxing Wang and Tianyi Gui and Tingyu Weng and Tong Shen and Wei Lin and Wei Wang and Wei Wang and Wenmeng Zhou and Wente Wang and Wenting Shen and Wenyuan Yu and Xianzhong Shi and Xiaoming Huang and Xin Xu and Yan Kou and Yangyu Lv and Yifei Li and Yijing Liu and Yiming Wang and Yingya Zhang and Yitong Huang and Yong Li and You Wu and Yu Liu and Yulin Pan and Yun Zheng and Yuntao Hong and Yupeng Shi and Yutong Feng and Zeyinzi Jiang and Zhen Han and Zhi-Fan Wu and Ziyu Liu},
      journal = {arXiv preprint arXiv:2503.20314},
      year={2025}
}

@article{hunyuanvideo,
  title={Hunyuanvideo: A systematic framework for large video generative models},
  author={Kong, Weijie and Tian, Qi and Zhang, Zijian and Min, Rox and Dai, Zuozhuo and Zhou, Jin and Xiong, Jiangfeng and Li, Xin and Wu, Bo and Zhang, Jianwei and others},
  journal={arXiv preprint arXiv:2412.03603},
  year={2024}
}

@inproceedings{lpips,
  title={The unreasonable effectiveness of deep features as a perceptual metric},
  author={Zhang, Richard and Isola, Phillip and Efros, Alexei A and Shechtman, Eli and Wang, Oliver},
  booktitle={2018 IEEE/CVF conference on computer vision and pattern recognition},
  pages={586--595},
  year={2018},
  organization={IEEE}
}

@article{ssim,
  title={{Image Quality Assessment: From Error Visibility to Structural Similarity}},
  author={Wang, Zhou and Bovik, Alan C. and Sheikh, Hamid R. and Simoncelli, Eero P.},
  journal={{IEEE Transactions on Image Processing}},
  volume={13},
  number={4},
  pages={600--612},
  year={2004},
  publisher={IEEE}
}

@article{imagereward,
  title={Imagereward: Learning and evaluating human preferences for text-to-image generation},
  author={Xu, Jiazheng and Liu, Xiao and Wu, Yuchen and Tong, Yuxuan and Li, Qinkai and Ding, Ming and Tang, Jie and Dong, Yuxiao},
  journal={Advances in Neural Information Processing Systems},
  volume={36},
  pages={15903--15935},
  year={2023}
}

@inproceedings{clipscore,
  title={Clipscore: A reference-free evaluation metric for image captioning},
  author={Hessel, Jack and Holtzman, Ari and Forbes, Maxwell and Le Bras, Ronan and Choi, Yejin},
  booktitle={Proceedings of the 2021 conference on empirical methods in natural language processing},
  pages={7514--7528},
  year={2021}
}

@article{drawbench,
  title={Photorealistic text-to-image diffusion models with deep language understanding},
  author={Saharia, Chitwan and Chan, William and Saxena, Saurabh and Li, Lala and Whang, Jay and Denton, Emily L and Ghasemipour, Kamyar and Gontijo Lopes, Raphael and Karagol Ayan, Burcu and Salimans, Tim and others},
  journal={Advances in neural information processing systems},
  volume={35},
  pages={36479--36494},
  year={2022}
}

@inproceedings{vbench,
  title={Vbench: Comprehensive benchmark suite for video generative models},
  author={Huang, Ziqi and He, Yinan and Yu, Jiashuo and Zhang, Fan and Si, Chenyang and Jiang, Yuming and Zhang, Yuanhan and Wu, Tianxing and Jin, Qingyang and Chanpaisit, Nattapol and others},
  booktitle={2024 IEEE/CVF Conference on Computer Vision and Pattern Recognition (CVPR)},
  pages={21807--21818},
  year={2024},
  organization={IEEE}
}

@article{cfg,
  title={Classifier-free diffusion guidance},
  author={Ho, Jonathan and Salimans, Tim},
  journal={arXiv preprint arXiv:2207.12598},
  year={2022}
}
\bibliographystyle{iclr2027_conference}

\appendix
\newpage
\section*{\Large Appendix}


\section{Kalman Filter}
\label{sec:bg_kalman}

The Kalman filter \citep{kalman1960} is the exact Bayesian estimator for a linear-Gaussian state-space model. A hidden state $h_j$ evolves linearly and is observed with noise,
\begin{equation}
  h_j = A_j h_{j-1} + w_j,\quad w_j \sim \mathcal{N}(0,Q_j),
  \qquad
  y_j = H_j h_j + \eta_j,\quad \eta_j \sim \mathcal{N}(0,R_j).
  \label{eq:kf}
\end{equation}
The filter represents its estimate of $h_j$ as a Gaussian with mean $\hat h_j$ and covariance $P_j$, and alternates between two operations. Between observations, it predicts the next state, $\hat h_j \leftarrow A_j \hat h_{j-1}$ and $P_j \leftarrow A_j P_{j-1} A_j^\top + Q_j$, so the estimate moves forward and its uncertainty grows. When an observation is available, it updates the prediction using the gain $K_j = P_j H_j^\top (H_j P_j H_j^\top + R_j)^{-1}$,
\begin{equation}
  \hat h_j \leftarrow \hat h_j + K_j \underbrace{(y_j - H_j \hat h_j)}_{\text{prediction residual}},
  \qquad
  P_j \leftarrow (I - K_j H_j)\,P_j .
  \label{eq:kf-update}
\end{equation}
The prediction residual is the part of the observation that the prediction did not account for. Under the model, its variance is $H_j P_j H_j^\top + R_j$. Comparing the observed residuals with this variance shows whether the assumed noise levels are too low or too high, which allows the filter to adjust them during inference.

For a smooth signal whose rate of change is unknown and varies slowly, a common choice of state is the \emph{local linear trend}, $h = (m, \nu)^\top$, where $m$ is the value and $\nu$ is its rate of change. The value integrates the rate, and the rate is perturbed by white noise. Both the transition and the process noise then depend only on the time $\Delta$ elapsed between steps,
\begin{equation}
  A(\Delta) = \begin{pmatrix} 1 & \Delta \\ 0 & 1 \end{pmatrix},
  \qquad
  Q(\Delta) = q \begin{pmatrix} \Delta^3/3 & \Delta^2/2 \\[2pt] \Delta^2/2 & \Delta \end{pmatrix},
  \label{eq:llt}
\end{equation}
where $Q(\Delta)$ is the covariance accumulated when the rate is perturbed by white noise of intensity $q$. SpecKCast uses this model with an additional damping factor on the rate (\cref{sec:method}).

This model has three properties that make it suitable for the forecasting problem in \eqref{eq:forecast}. First, the predicted value is the minimum-mean-squared-error estimate under the model and depends on all past observations. When the rate is poorly determined, the gain reduces its influence on the forecast. A finite-difference forecaster instead relies only on the most recent observations and amplifies their noise. Second, the filter handles irregularly spaced observations directly, because $\Delta$ enters $A$ and $Q$. Fixed-order difference formulas assume uniform spacing, which an adaptive compute schedule does not provide. Third, the predicted covariance $P$ is available before the next observation, so the filter estimates its own forecast error. This lets us choose the compute steps $\mathcal{U}$ by thresholding the predicted error rather than with a hand-designed schedule.

The Kalman filter is related to the Wiener filter, which is the optimal linear filter for a stationary signal in stationary noise and is defined by a fixed frequency response. The Kalman filter is its recursive counterpart for non-stationary signals: it tracks an evolving state together with its uncertainty. The filter itself is standard. Our contribution is the state-space model we place on the denoiser output and the way its covariance is calibrated and used for scheduling (\cref{sec:method}).



\section{Details of the Band-specific Time Warp}
\label{app:warp}

This appendix expands on the time variable introduced in \cref{sec:warp}. We derive the interpretation of $\lambda_j + \log p_\ell$ as the band's own signal-to-noise ratio, describe the mechanics of the warp, and collect the implementation details omitted from the main text.

\subsection{The Warped Quantity is the Band's Own Log-SNR}
\label{app:warp_snr}

Recall that a noisy latent is a linear interpolation between data and noise, $x_\sigma = (1-\sigma)\,x_0 + \sigma\,\epsilon$ with $\epsilon \sim \mathcal{N}(0, I)$. Because the DCT is orthonormal, transforming white Gaussian noise yields white Gaussian noise, so every frequency coefficient of the noise term has unit variance and the noise contributes power $\sigma^2$ uniformly to every band. The data term is not uniform. Latents of natural images concentrate energy at low frequencies, so the coefficients of band $\ell$ carry an average signal power $p_\ell$ that decreases with frequency. The signal-to-noise ratio of band $\ell$ at noise level $\sigma_j$ is therefore approximately
\begin{equation}
\mathrm{SNR}_\ell(\sigma_j)
\;=\;
\frac{(1-\sigma_j)^2\, p_\ell}{\sigma_j^2}
\qquad\Longrightarrow\qquad
\log \mathrm{SNR}_\ell(\sigma_j)
\;=\;
\underbrace{2\log\!\big(\tfrac{1-\sigma_j}{\sigma_j}\big)}_{\lambda_j}
\;+\; \log p_\ell .
\label{eq:band_snr}
\end{equation}
The sigmoid in~\eqref{eq:warp} is therefore applied to the band's own log-SNR, scaled by $1/T_w$. Its argument crosses zero when $\mathrm{SNR}_\ell = 1$, the point at which the band's signal emerges from the noise. Since $p_\ell$ is large for low bands and small for high bands, low bands cross this threshold early in denoising (at large $\sigma_j$) and high bands cross it late, which explains why different bands do most of their changing in different phases.

\subsection{Mechanics of the Warp}
\label{app:warp_mechanics}

The sigmoid maps the band's scaled log-SNR to $(0,1)$, and its derivative is largest at the zero crossing and decays symmetrically away from it. Consequently, most of the increase in the warped time $t_\ell(j) = \operatorname{sigmoid}\big((\lambda_j + \log p_\ell)/T_w\big)$ occurs during the steps around the band's emergence, and $t_\ell$ barely moves while the band is still dominated by noise or has already settled. The temperature $T_w$ sets the width of this active window in log-SNR units; a small $T_w$ concentrates the time variable tightly around the crossing, while a large $T_w$ approaches a plain rescaling of $\lambda_j$. We use $T_w = 2$ for all bands and all models.

Since $\sigma_j$ decreases monotonically over sampling, $\lambda_j$ increases monotonically, and both the sigmoid and the final rescaling are monotone maps. Each $s_\ell$ is therefore a strictly increasing function of the step index, a requirement for a valid time variable. The operator $\operatorname{rescale}_{[-1,1]}$ is the affine map taking the smallest and largest values of $t_\ell$ over the $N$ steps of the run to $-1$ and $1$, respectively. Since $t_\ell$ is increasing, these are its first and last values, so
\begin{equation}
s_\ell(j) = 2\,\frac{t_\ell(j) - t_\ell(0)}{t_\ell(N-1) - t_\ell(0)} - 1 .
\label{eq:rescale}
\end{equation}
Every band's time variable therefore spans the same interval regardless of where its active window lies.

The warp interacts with the forecaster through the increments $\Delta_{\ell,j} = s_\ell(j) - s_\ell(j-1)$, which serve as the effective step sizes of the per-band Kalman filters. During a band's active window the increments are large, so the filter expects more change per step, injects more process noise, and damps its trend estimate faster; outside the window the increments are small and the filter treats the band as nearly static. The warp thus lets a single, simple dynamics model adapt its behavior to each band and phase without any per-step logic.



\subsection{Implementation Details}
\label{app:warp_impl}

The following details matter for reproducing the coordinate exactly. First, rectified-flow pipelines typically apply a shift to the noise schedule, which for some models depends on the resolution, so the realized $\sigma_j$ differ from the nominal linearly spaced values. SpecKCast rebuilds the exact schedule the pipeline will use before sampling begins and computes all $s_\ell(j)$ from it, so the forecaster's time coordinates line up with the transformer calls at any resolution and step count. Second, the endpoints of the schedule are clamped away from $\sigma = 0$ and $\sigma = 1$ before evaluating $\lambda_j$ to avoid infinities. The clamping changes only the first and last values of $\lambda_j$, where the sigmoid is nearly flat, so its effect on $s_\ell$ is small. Third, the band powers $p_\ell$ are measured once during calibration (\cref{sec:calib}) from the implied clean latent $\hat x_0 = x_j - \sigma_j Y_j$ over the final steps of the calibration trajectories. Since the warp is a fixed elementwise function of the schedule, the entire coordinate system costs a few scalar operations per run and is computed once before sampling.

\subsection{Band-Power Calibration}
\label{sec:calib}

SpecKCast uses $p_\ell$ to account for differences in signal magnitude across frequency bands. We estimate these values once per model from a small calibration set $\mathcal C$ by running the unaccelerated model and recording the latent $x_j$, model output $Y_j$, and noise level $\sigma_j$ at every denoising step.

Under the rectified-flow interpolation,
\begin{equation}
x_j
=
(1-\sigma_j)x_0+\sigma_j\epsilon
=
x_0+\sigma_j(\epsilon-x_0).
\end{equation}
Because the model predicts the velocity $Y_j\approx\epsilon-x_0$, it provides the clean-sample estimate
\begin{equation}
\widehat{x}_{0,j}=x_j-\sigma_jY_j.
\label{eq:x0_estimate}
\end{equation}
We estimate power using the later denoising steps, where $\sigma_j$ is smaller. In particular, if
\begin{equation}
Y_j=(\epsilon-x_0)+\delta_j,
\end{equation}
then
\begin{equation}
\widehat{x}_{0,j}=x_0-\sigma_j\delta_j,
\end{equation}
so velocity-prediction errors have less influence on $\widehat{x}_{0,j}$ near the end of denoising.

We apply the same channel-wise orthonormal DCT used by SpecKCast,
\begin{equation}
X_{0,j}
=
\operatorname{DCT}_{2\mathrm{D}}(\widehat{x}_{0,j}).
\end{equation}
Let $\Omega_\ell$ contain the spatial-frequency indices belonging to band $\ell$. We estimate its signal power by averaging the squared DCT coefficients over calibration samples, selected denoising steps, channels, and frequencies:
\begin{equation}
p_\ell
=
\frac{1}
{|\mathcal C||\mathcal J|C|\Omega_\ell|}
\sum_{n\in\mathcal C}
\sum_{j\in\mathcal J}
\sum_{c=1}^{C}
\sum_{(u,v)\in\Omega_\ell}
\left|X_{0,j,c,u,v}^{(n)}\right|^2,
\label{eq:band_power}
\end{equation}
where $\mathcal J$ denotes the selected denoising steps.

As shown in \cref{app:warp_snr}, $\lambda_j + \log p_\ell$ is the log-SNR of band $\ell$. Thus, $p_\ell$ shifts the global log-SNR according to the signal strength of each band, so the resulting coordinate reflects when that band transitions from noise-dominated to signal-dominated.

In our implementation, the calibration uses 12 prompts with 50 denoising steps each, and $\mathcal J$ contains the final 25 steps of each trajectory. The resulting $p_\ell$ are fixed across prompts during inference. Calibration modifies no model parameters and introduces no additional model evaluations at inference time.

\paragraph{Calibration-free alternatives.}
Offline calibration is not required to apply SpecKCast. The simplest alternative is to disable the band-specific warp and use rescaled log-SNR for every band,
\begin{equation}
s_\ell(j)
=
\operatorname{rescale}_{[-1,1]}(\lambda_j).
\end{equation}
Forecasting remains valid, but all bands then share the same temporal coordinate. Consequently, elapsed time no longer reflects their different signal-to-noise transitions, removing the band-specific control of extrapolation, damping, and uncertainty. This variant corresponds to the raw-logSNR row of \Cref{tab:ablation-main}, where it lowers PSNR by 3.3\,dB relative to the band-specific warp.

Alternatively, $p_\ell$ can be estimated online from the first $K$ computed outputs,
\begin{equation}
p_\ell
=
\frac{1}{K d_\ell}
\sum_{j=0}^{K-1}
\sum_{i\in\mathcal B_\ell}
Z_{i,j}^{2},
\end{equation}
where $\mathcal B_\ell$ contains the coefficients of band $\ell$ across all channels and $d_\ell=|\mathcal B_\ell|=C|\Omega_\ell|$. When these outputs come from the warm-up steps (\cref{app:sched_rule}), this approach requires no additional transformer evaluations. However, the warm-up steps occur at the start of sampling, when the input is almost pure noise. The model output at these steps is therefore close to the noise itself, which has the same power in every band. As a result, the estimated $p_\ell$ is nearly the same for all bands and does not capture how signal power decreases with frequency. For this reason, all our experiments use the offline calibration described above.

\subsection{Rationale for Band-Specific Time Reparameterization}
\label{app:warp}

The denoising process follows one global sequence of steps $j$ and noise levels $\sigma_j$. SpecKCast does not assign different physical denoising times to different frequency bands. Instead, it represents the same global step using a different increasing coordinate for each band:
\begin{equation}
s_\ell(j)=g_\ell(\lambda_j),
\label{eq:band_reparameterization}
\end{equation}
where $g_\ell$ is the warp associated with band $\ell$. Because each $g_\ell$ is strictly increasing, the ordering of the denoising steps is preserved:
\begin{equation}
j_1<j_2
\quad\Longrightarrow\quad
s_\ell(j_1)<s_\ell(j_2).
\end{equation}
Thus, every band follows the same denoising direction and reaches the same target step, although the effective distance between two steps can differ across bands.
For a skip from step $j$ to step $k$, band $\ell$ experiences the effective time interval
\begin{equation}
\Delta_\ell(j,k)
=
s_\ell(k)-s_\ell(j).
\label{eq:band_time_interval}
\end{equation}
This interval can be large for a band expected to change substantially during the skip and small for a band expected to remain stable. Consequently, the forecaster can make a larger update for an active band and a more conservative prediction for a stable one. This principle is not specific to the Kalman filter. Any forecasting model that uses time as an input, including linear extrapolation and polynomial fitting, can operate in a band-specific coordinate.
The warp used by SpecKCast is
\begin{equation}
g_\ell(\lambda)
=
\operatorname{rescale}_{[-1,1]}
\left[
\operatorname{sigmoid}
\left(
\frac{\lambda+\log p_\ell}{T_w}
\right)
\right].
\end{equation}
For $p_\ell>0$ and $T_w>0$, the sigmoid term is strictly increasing in $\lambda$, and the affine rescaling preserves this. The rescaling also gives all bands the same endpoints $-1$ and $1$, while their progression between these endpoints can differ.


\section{Details of the Kalman Forecaster}
\label{app:kalman_matrices}

The transition matrix $A_{\ell,j}$ and process-noise covariance $Q_{\ell,j}$ serve different purposes. The former describes the expected evolution of a coefficient, while the latter represents deviations from that evolution. For coefficient $i$ in band $\ell$, the state is
\begin{equation}
h_{i,j}=(m_{i,j},\nu_{i,j}),
\end{equation}
where $m_{i,j}$ is the coefficient value and $\nu_{i,j}$ is its temporal derivative. Over the interval $\Delta_{\ell,j}$, we advance the value using its derivative and attenuate the derivative by
\begin{equation}
\rho_{\ell,j}
=
\exp(-\Delta_{\ell,j}/\tau_\ell).
\end{equation}
These operations give
\begin{equation}
A_{\ell,j}
=
\begin{bmatrix}
1 & \Delta_{\ell,j}\\
0 & \rho_{\ell,j}
\end{bmatrix}.
\end{equation}
The upper-right entry integrates the derivative over the interval, while the lower-right entry prevents old changes from being extrapolated indefinitely. As the interval grows, $\rho_{\ell,j}$ decreases and the prediction relies less on the previous derivative.

The deterministic transition cannot capture all changes in the coefficient trajectory. We model these deviations as continuous random perturbations to the temporal derivative. If their intensity is $q_{\ell,j}$ and the damping is ignored, the accumulated state perturbation over an interval $\Delta_{\ell,j}$ has covariance
\begin{equation}
Q_{\ell,j}
=
q_{\ell,j}
\int_0^{\Delta_{\ell,j}}
\begin{bmatrix}
(\Delta_{\ell,j}-u)^2 & \Delta_{\ell,j}-u\\
\Delta_{\ell,j}-u & 1
\end{bmatrix}
\,du,
\end{equation}
which evaluates to
\begin{equation}
Q_{\ell,j}
=
q_{\ell,j}
\begin{bmatrix}
\Delta_{\ell,j}^3/3 & \Delta_{\ell,j}^2/2\\
\Delta_{\ell,j}^2/2 & \Delta_{\ell,j}
\end{bmatrix}.
\label{eq:process_noise_derivation}
\end{equation}
The lower-right term describes uncertainty accumulated in the temporal derivative. Integrating that uncertainty produces the cross-covariance terms and the cubic growth of uncertainty in the coefficient value. Consequently, longer forecast intervals result in larger covariance.

The scalar $q_{\ell,j}$ controls the magnitude of this uncertainty. Bands with greater temporal activity receive larger process intensity, allowing their observations to produce stronger corrections. Since coefficients in the same band share $\Delta_{\ell,j}$, $\rho_{\ell,j}$, $q_{\ell,j}$, and the observation noise, they also share $A_{\ell,j}$, $Q_{\ell,j}$, and the resulting Kalman covariance, while retaining separate state estimates.

This construction has intuitive limiting behavior. As $\Delta_{\ell,j}\rightarrow0$, $A_{\ell,j}\rightarrow I$ and $Q_{\ell,j}\rightarrow0$, since little evolution or uncertainty is expected over a negligible interval. For longer intervals, the derivative is increasingly attenuated while process uncertainty grows.

This transition is a stepwise discrete approximation rather than the exact discretization of a continuous damped system. It first advances the coefficient using the derivative at the beginning of each interval and then attenuates that derivative. Because the prediction is applied successively over all intervening denoising steps, the damping compounds across a multi-step skip.

\paragraph{Bounded accumulation across skipped steps.}
The damping acts through repeated prediction over the intervening denoising steps, rather than within a single large prediction. To illustrate this, consider $K$ skipped steps with a constant increment $\Delta_\ell$, so that $\rho_\ell$ is also constant. Indexing steps from the start of the skip, the temporal derivative after $k$ steps is
\begin{equation}
\nu_{i,k}
=
\rho_\ell^k\nu_{i,0}.
\end{equation}
The total change in the predicted coefficient value is therefore
\begin{equation}
m_{i,K}-m_{i,0}
=
\sum_{k=0}^{K-1}
\Delta_\ell\rho_\ell^k\nu_{i,0}
=
\Delta_\ell
\frac{1-\rho_\ell^K}{1-\rho_\ell}
\nu_{i,0}.
\label{eq:bounded_extrapolation}
\end{equation}
Consequently,
\begin{equation}
\left|m_{i,K}-m_{i,0}\right|
\leq
\frac{\Delta_\ell}{1-\rho_\ell}
\left|\nu_{i,0}\right|.
\end{equation}
Thus, the contribution of the initial temporal derivative remains bounded as the skip length increases. In contrast, without damping, $\rho_\ell=1$ and the displacement grows linearly as $K\Delta_\ell\nu_{i,0}$, limited only by the length of the coordinate interval. Moreover,
\begin{equation}
\lim_{\Delta_\ell\rightarrow0}
\frac{\Delta_\ell}
{1-\exp(-\Delta_\ell/\tau_\ell)}
=
\tau_\ell,
\end{equation}
so the bound approaches $\tau_\ell|\nu_{i,0}|$ as the discretization is refined. Without damping, the displacement can reach $2|\nu_{i,0}|$, since $s_\ell$ spans $[-1,1]$. 

For nonuniform increments, the derivative attenuation still composes exactly:
\begin{equation}
\prod_{r=1}^{K}
\rho_{\ell,r}
=
\exp\left(
-\frac{\sum_{r=1}^{K}\Delta_{\ell,r}}{\tau_\ell}
\right).
\end{equation}
The accumulated change in value depends on the underlying denoising grid, but the derivative continues to decay with the total elapsed band time. This bound concerns the deterministic extrapolation induced by the current derivative; it does not by itself bound the total forecasting error.

\paragraph{Process-noise covariance.}
\Cref{eq:process_noise_derivation} corresponds to adding white noise to the temporal derivative while ignoring damping,
\begin{equation}
d m = \nu\,ds,
\qquad
d\nu = \sqrt{q_{\ell,j}}\,dW_s,
\end{equation}
where $W_s$ is a Wiener process. This is the standard integrated-Wiener covariance (\cref{sec:bg_kalman}). The resulting matrix is positive definite for $q_{\ell,j}>0$ and $\Delta_{\ell,j}>0$, since
\begin{equation}
\det(Q_{\ell,j})
=
\frac{q_{\ell,j}^2\Delta_{\ell,j}^4}{12}>0.
\end{equation}

We deliberately combine this undamped covariance with the stepwise damped transition as a simple uncertainty approximation. To justify this choice, consider the exact continuous damped model with $\rho_{\ell,j}=\exp(-\Delta_{\ell,j}/\tau_\ell)$. Its diagonal process variances are
\begin{align}
Q_{11}^{\mathrm{damp}}
&=
q_{\ell,j}\tau_\ell^2
\left[
\Delta_{\ell,j}
-2\tau_\ell(1-\rho_{\ell,j})
+\frac{\tau_\ell}{2}(1-\rho_{\ell,j}^2)
\right],\\
Q_{22}^{\mathrm{damp}}
&=
\frac{q_{\ell,j}\tau_\ell}{2}
(1-\rho_{\ell,j}^2).
\end{align}
Applying
\begin{equation}
\tau_\ell(1-e^{-x/\tau_\ell})\leq x,
\qquad
e^{-x/\tau_\ell}\leq1
\end{equation}
inside the integrals that define these variances gives
\begin{equation}
Q_{11}^{\mathrm{damp}}
\leq
q_{\ell,j}\frac{\Delta_{\ell,j}^3}{3},
\qquad
Q_{22}^{\mathrm{damp}}
\leq
q_{\ell,j}\Delta_{\ell,j}.
\end{equation}
Our choice therefore overestimates the marginal variance growth relative to the exact damped model. This is the conservative direction for uncertainty-based scheduling, since it favors earlier model evaluation rather than overconfident forecasting. For small adjacent intervals, the damped and undamped forms agree to leading order. Any remaining mismatch is absorbed by the calibrated and online-adapted process intensity $q_{\ell,j}$.



\section{Details of the Uncertainty-Aware Adaptive Scheduler}
\label{app:scheduling}

This appendix specifies the details behind \Cref{sec:scheduling}: the covariance recursion used to evaluate the risk, its calibrated noise and online correction, and the exact decision rule with all safeguards.

\subsection{The Risk Sidecar}
\label{app:sched_sidecar}

The risk of \Cref{eq:risk} is evaluated by a dedicated covariance recursion, which we call the \emph{sidecar}, that runs alongside the mean filter of \Cref{sec:kalman} but is decoupled from it in two ways. First, it operates in \emph{relative} units, tracking forecast error as a fraction of each band's power, which makes its outputs comparable across bands and generations. Second, its process noise is not the dynamics noise of the mean filter but a table $Q^{\mathrm{sc}}_{\ell,w}$ calibrated directly against the forecaster's realized errors, so the sidecar estimates how wrong the forecast tends to be after a given number of skipped steps. The index $w$ ranges over four step windows partitioning the schedule, allowing the noise to differ between early, middle, and late denoising.

The sidecar maintains one $2 \times 2$ covariance per band with entries $(\tilde p_{11}, \tilde p_{12}, \tilde p_{22})$ and propagates them with the same integrated-noise structure as
\begin{equation}
\begin{aligned}
\tilde p_{11} &\leftarrow \tilde p_{11} + 2\Delta_\ell\, \tilde p_{12} + \Delta_\ell^{2}\, \tilde p_{22} + Q^{\mathrm{sc}}_{\ell,w}\, \Delta_\ell^{3}/3, \\
\tilde p_{12} &\leftarrow \rho_{\mathrm{sc}} \big(\tilde p_{12} + \Delta_\ell\, \tilde p_{22}\big) + Q^{\mathrm{sc}}_{\ell,w}\, \Delta_\ell^{2}/2, \\
\tilde p_{22} &\leftarrow \rho_{\mathrm{sc}}^{2}\, \tilde p_{22} + Q^{\mathrm{sc}}_{\ell,w}\, \Delta_\ell,
\end{aligned}
\label{eq:app_sidecar}
\end{equation}
with a fixed memory factor $\rho_{\mathrm{sc}} = 0.9$ per step. At every computed step, the sidecar conditions on the fact that the value has just been observed exactly. The trend uncertainty is reduced to its posterior, and the value uncertainty and cross term collapse to zero,
\begin{equation}
\tilde p_{22} \leftarrow \tilde p_{22} - \tilde p_{12}^{2} / \tilde p_{11},
\qquad
\tilde p_{11} \leftarrow 0,
\qquad
\tilde p_{12} \leftarrow 0,
\label{eq:app_collapse}
\end{equation}
so the risk restarts from zero after every transformer call and re-accumulates over the following skips. The quantity $P^{(\ell)}_{11}$ in \Cref{eq:risk} refers to $\tilde p_{11}$ propagated by \Cref{eq:app_sidecar}.

Like the mean filter, the sidecar corrects itself online. At each computed step $j$, we compare the observed forecast error of band $\ell$ with the sidecar's predicted variance,
\begin{equation}
r_{\ell,j}
=
\frac{1}{d_\ell\, p_\ell}
\sum_{i\in\mathcal B_\ell}
\big(Z_{i,j}-\bar m_{i,j}\big)^2,
\qquad
\tilde\kappa_\ell
\leftarrow
\operatorname{clip}_{[0.1,\,10]}
\Big(
0.8\,\tilde\kappa_\ell + 0.2\,\frac{r_{\ell,j}}{\tilde p^{(\ell)}_{11}(j)}
\Big),
\label{eq:app_kappa}
\end{equation}
where $\bar m_{i,j}$ is the predicted coefficient value before the Kalman update, $\tilde p^{(\ell)}_{11}(j)$ is the sidecar variance before it is reset by \Cref{eq:app_collapse}, and $\tilde\kappa_\ell$ is initialized to $1$. The scale $\tilde\kappa_\ell$ multiplies the band's contribution to the risk. If a generation's observed errors are larger than the sidecar predicted, the ratio exceeds 1 and $\tilde\kappa_\ell$ increases. The risk then reaches the threshold sooner, so the transformer is evaluated more often and skips become shorter. If the observed errors are smaller than predicted, $\tilde\kappa_\ell$ decreases and the scheduler allows longer skips. The full risk is
\begin{equation}
R(j) \;=\; \omega(j) \sum_{\ell} \tilde\kappa_\ell\; \gamma_{\ell, w(j)}\; \tilde p^{(\ell)}_{11}(j),
\label{eq:app_risk}
\end{equation}
which reduces to \Cref{eq:risk} when $\tilde\kappa_\ell = 1$ and $\gamma_{\ell,w} = d_\ell$. The perceptual weights $\gamma_{\ell,w}$ are calibrated by perturbation, injecting band-limited errors into the output at different phases of sampling and measuring the LPIPS change of the final sample. If these weights have not been calibrated, we use $\gamma_{\ell,w} = d_\ell$, the number of coefficients in band $\ell$. The risk then measures the total expected error summed over all coefficients, giving every coefficient the same importance.

\subsection{Decision Rule}
\label{app:sched_rule}

After each computed step $j$, the sidecar is rolled forward \emph{predictively}, applying \Cref{eq:app_sidecar} over a horizon of up to $10$ future steps without changing its state, and the resulting risks $R(j{+}1), R(j{+}2), \dots$ are compared against the threshold. The next computed step is placed at the first predicted crossing $R > \varepsilon$; all steps before it are forecast. Three safeguards apply. The first $3$ steps of sampling are always computed, giving the filters real observations to establish values and trends; the final step is always computed, so the sample is finished by a true model evaluation; and the number of consecutive skips is capped at $10$ regardless of the predicted risk. The threshold $\varepsilon$ is set once for each compute budget. Starting from a range of candidate values, we run the scheduler on a set of calibration prompts that are not used for evaluation and measure the average number of transformer calls. If the average is above the target, we raise $\varepsilon$; if it is below, we lower it. We repeat this, halving the range each time, until the average matches the target.

At inference, $\varepsilon$ is kept fixed, but the number of transformer calls is not. The risk depends on the per-band scale $\tilde\kappa_\ell$, which is updated from the errors observed during each generation. A generation whose trajectory is easy to forecast accumulates risk slowly and uses fewer calls than the target, while a harder one uses more. The target is therefore met on average rather than for every generation, and the number of calls varies around the target from one generation to the next.

\subsection{Fixed-Schedule Variant}
\label{app:sched_fixed}

As a deterministic alternative, we support a fixed schedule with linearly growing gaps, computing steps
\begin{equation}
j_r = \big\lfloor \alpha\, r(r{+}1)/2 \big\rfloor, \qquad r = 1, 2, \dots, \quad j_r < N-1,
\label{eq:app_fixed}
\end{equation}
together with the warm-up steps $\{0,1,2\}$ and the final step, where $\alpha$ is chosen by a simple search so that the total number of computed steps matches the target budget. The spacing is dense early in sampling, where the output changes fastest, and sparse late, similar to the schedules the adaptive scheduler produces for typical generations. The fixed variant is preferable when a deterministic compute cost is required, while the adaptive variant reallocates calls between easy and hard generations at the same average cost. In \Cref{tab:ablation-main}, the adaptive variant improves PSNR by 2.2\,dB over the fixed variant at the same budget.


\clearpage
\section{Models with Classifier-Free Guidance}
\label{app:cfg}

Some of the evaluated models use classifier-free guidance (CFG)~\citep{cfg} at inference. For these models, each denoising step runs the transformer twice, once with the prompt and once with the negative prompt, and combines the conditional output $Y^{c}_j$ and the unconditional output $Y^{u}_j$ into a guided velocity
\begin{equation}
Y_j = \mathcal{G}\big(Y^{c}_j, Y^{u}_j\big),
\qquad
\mathcal{G}(c,u) = u + w\,(c-u),
\label{eq:app_cfg}
\end{equation}
where $w$ is the guidance scale. Some models modify $\mathcal{G}$; Qwen-Image, for example, rescales each token of the guided velocity to the norm of the conditional prediction (\cref{eq:app_qwen_guidance}).

\paragraph{Forecasting the guided velocity.}
SpecKCast forecasts the transformer output rather than its internal features, so it can be placed outside the network. For models with CFG, we place it after the guidance combination and forecast the guided velocity $Y_j$, which is the quantity the sampler uses, with a single forecaster. On a computed step, both passes run, $Y_j$ is formed with the model's own $\mathcal{G}$, and the forecaster is updated with it. On a skipped step, neither pass runs and the forecast $\widehat{Y}_j$ is passed to the sampler. The frequency bands, time coordinates, Kalman filters, and scheduler are the same as for models without guidance, which correspond to $\mathcal{G}(c,u) = c$. With $n$ computed steps out of $N$, SpecKCast therefore uses $2n$ transformer passes instead of $2N$.

\paragraph{Relation to forecasting each pass separately.}
An alternative is to forecast $Y^{c}_j$ and $Y^{u}_j$ separately on the same schedule and combine the two forecasts with $\mathcal{G}$. Suppose the forecast at a skipped step $k$ is a linear combination of the outputs observed at earlier computed steps,
\begin{equation}
\widehat{Y}_k = \sum_{j \in \mathcal{U},\, j<k} a_{k,j}\, Y_j,
\label{eq:app_linear_forecast}
\end{equation}
where the weights $a_{k,j}$ do not depend on the observed values. If, in addition, $\mathcal{G}(c,u) = w\,c + (1-w)\,u$ with the same $w$ at every step, then
\begin{equation}
\sum_{j} a_{k,j}\,\mathcal{G}\big(Y^{c}_j, Y^{u}_j\big)
=
w \sum_{j} a_{k,j}\, Y^{c}_j + (1-w) \sum_{j} a_{k,j}\, Y^{u}_j
=
\mathcal{G}\big(\widehat{Y}^{c}_k, \widehat{Y}^{u}_k\big),
\label{eq:app_cfg_equivalence}
\end{equation}
so forecasting the guided velocity gives the same output as forecasting each pass and combining the forecasts. Taylor extrapolation and the Chebyshev fit used by Spectrum have the form of \eqref{eq:app_linear_forecast}, and so does a Kalman filter with fixed noise parameters, since its gains do not depend on the observed values. In SpecKCast, the equivalence is no longer exact because the process noise is adapted online from the observed errors, which makes the weights $a_{k,j}$ depend on the observations. It is also not exact when $\mathcal{G}$ includes a nonlinear correction such as the norm rescaling of Qwen-Image. We forecast the guided velocity because it is the quantity the sampler uses, so the forecast error and the scheduling risk are measured directly on what affects the sample, and because it requires a single set of filter states instead of two.

\paragraph{Variants.}
We also implement two variants, which are not used in the reported experiments. The per-branch variant runs one forecaster for each branch on a shared schedule, set by the conditional branch, and combines the two forecasts with $\mathcal{G}$. The asymmetric variant gives each branch its own schedule and compute budget, so a step can run neither pass, only one, or both. Writing \eqref{eq:app_cfg} as $Y_j = w\,Y^{c}_j - (w-1)\,Y^{u}_j$ shows that an error in the unconditional forecast is scaled by $w-1$, while an error in the conditional forecast is scaled by $w$. This motivates spending fewer evaluations on the unconditional branch.


\clearpage
\section{Extended Qualitative Comparisons}
\label{app:extended_comparisons}

We extend the comparisons in \Cref{sec:t2i,sec:t2v} with additional Qwen-Image, HunyuanVideo, and Wan2.1-14B examples. Each panel reports its evaluation count and measured speedup relative to the 50-step reference. These examples illustrate the visual effects of acceleration; aggregate results are reported in \Cref{tab:image-maintable,tab:video_main}.

\paragraph{Why SpecKCast preserves the reference more closely.}
As discussed in \Cref{sec:relatedwork}, cached features become outdated over long skips, while feature-space extrapolation does not distinguish the temporal dynamics of different spatial frequencies. SpecKCast forecasts the velocity passed to the sampler one DCT coefficient at a time, using band-specific time coordinates to account for these differences. Damped trends limit overshooting, fresh evaluations correct the forecast, and uncertainty-aware scheduling allocates computation when the predicted error grows. The ablations in \Cref{tab:ablation-main} support the contributions of frequency decomposition, time warping, and adaptive scheduling; the examples below illustrate the combined method.

\paragraph{Qwen-Image: color, texture, and object structure.}
In \Cref{fig:app_qwen_comparison}, TaylorSeer and HiCache give the strawberry scene a strong blue cast and darken the fox scene. TeaCache changes the basket's contents and introduces a second astronaut, while SeaCache adds an extra horse leg and smooths the fox's fur. Spectrum introduces grain and distorts the rider. SpecKCast more closely retains the reference lighting, object arrangement, and recognizable details, with $6.19$--$6.21\times$ speedup at 8 NFE and $4.97\times$ at 10 NFE. SpecKCast's closer match to the reference is visible in both overall appearance and local structure, at lower latency than the displayed baselines.

\begin{figure}[!htbp]
  \centering
  \includegraphics[width=\linewidth]{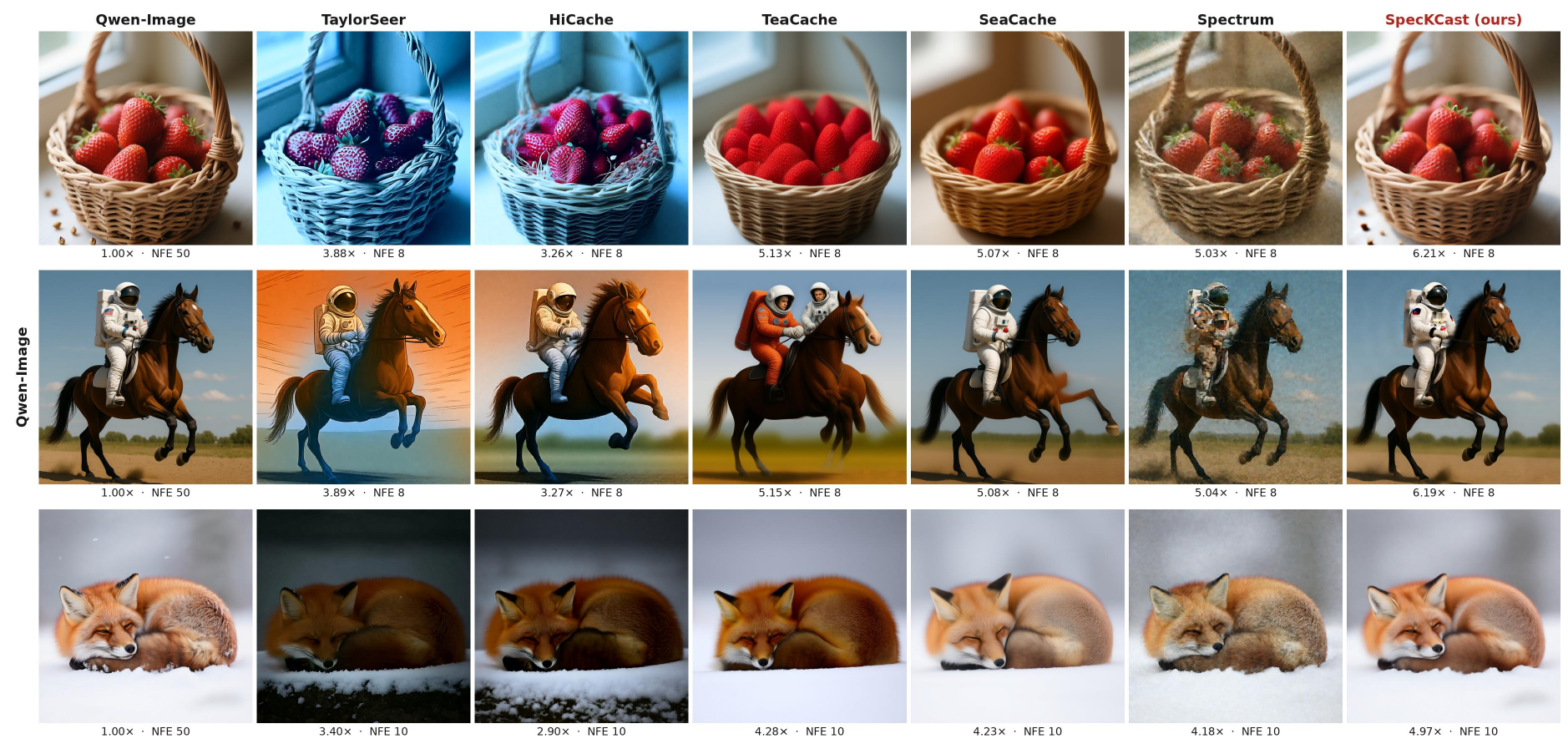}
  \caption{\textbf{Additional Qwen-Image comparisons.} Rows show strawberries in a basket, an astronaut riding a horse, and a fox in snow. The leftmost column is the 50-NFE reference; all accelerated methods use 8 NFE in the first two rows and 10 NFE in the last, with the measured speedup shown below each image. SpecKCast more closely preserves the reference colors and object structure, while the baselines exhibit color shifts, altered content, or degraded texture.}
  \label{fig:app_qwen_comparison}
\end{figure}

\clearpage
\paragraph{HunyuanVideo: subject detail and geometry.}
In \Cref{fig:app_hunyuan_bottle}, FORA and TaylorSeer blur the face, TeaCache changes the action and produces overlapping hand and face structure, and HiCache changes the subject's appearance. SpecKCast produces a clearer face and retains the bottle in hand across the displayed frames at 10 NFE and $4.27\times$ speedup, although the subject's appearance still differs from the reference. In \Cref{fig:app_hunyuan_bicycle}, TaylorSeer and Spectrum introduce an extra wheel-like structure between the bicycle's two wheels. SpecKCast preserves the two-wheel geometry with fewer evaluations (13 NFE compared with 14 and 16), though it omits the reference's front basket.

\begin{figure}[!htbp]
  \centering
  \includegraphics[width=\linewidth,height=0.70\textheight,keepaspectratio]{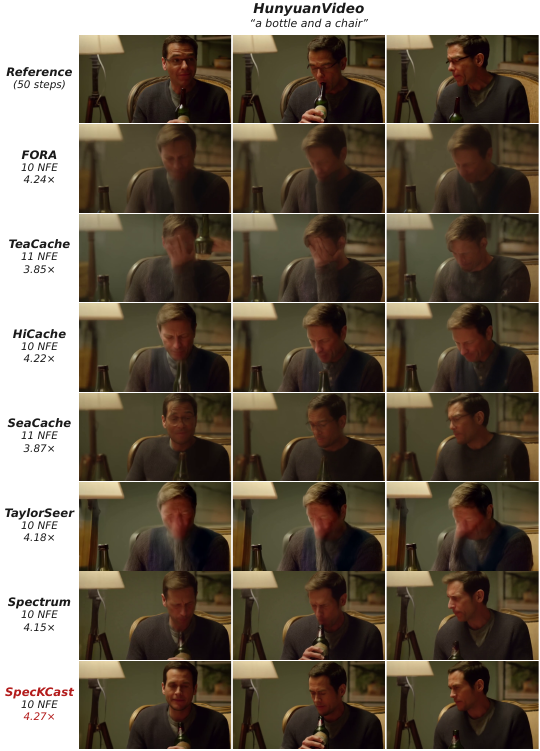}
  \caption{\textbf{HunyuanVideo: ``a bottle and a chair.''} Frames 0, 40, and 80 for each method show differences in facial detail, pose, and bottle placement. SpecKCast uses 10 NFE ($4.27\times$), compared with 10--11 NFE ($3.85$--$4.24\times$) for the baselines. It avoids the severe facial blur visible in FORA and TaylorSeer and keeps the bottle in hand, although the subject's appearance still differs from the reference.}
  \label{fig:app_hunyuan_bottle}
\end{figure}

\begin{figure}[!htbp]
  \centering
  \includegraphics[width=\linewidth]{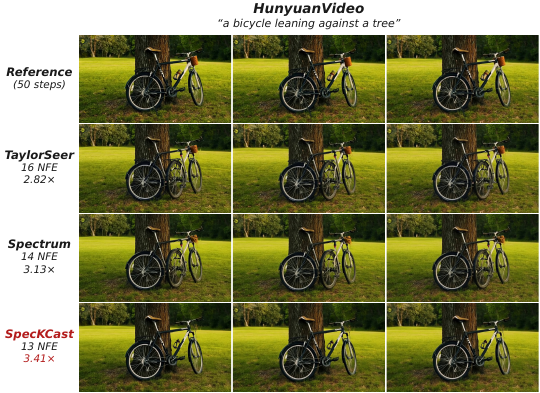}
  \caption{\textbf{HunyuanVideo: ``a bicycle leaning against a tree.''} TaylorSeer (16 NFE, $2.82\times$) and Spectrum (14 NFE, $3.13\times$) introduce an additional wheel-like structure between the bicycle's two wheels. SpecKCast (13 NFE, $3.41\times$) better preserves the main bicycle geometry across the displayed frames, although the front basket is missing.}
  \label{fig:app_hunyuan_bicycle}
\end{figure}

\paragraph{Wan2.1-14B: appearance and scene structure at 8 NFE.}
In \Cref{fig:app_wan_panda}, HiCache and TaylorSeer produce strong saturation and blurred boundaries. FORA and TeaCache retain the layout but soften the panda's detail, while SeaCache changes its appearance and the tableware at 7 NFE. SpecKCast retains the reference composition and drinking poses at $5.61\times$ speedup. In \Cref{fig:app_wan_boat}, TaylorSeer severely distorts the boat and skyline, and Spectrum alters the Eiffel Tower's shape. SpecKCast retains a more faithful landmark silhouette and river layout at the same 8 NFE and $5.65\times$ speedup as Spectrum.

\begin{figure}[!htbp]
  \centering
  \includegraphics[width=\linewidth]{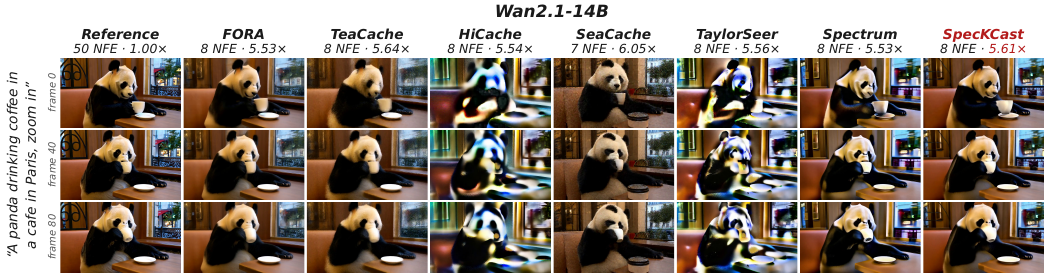}
  \caption{\textbf{Wan2.1-14B: a panda drinking coffee with a zoom-in.} Frames 0, 40, and 80 show subject appearance and scene layout. All accelerated methods use 8 NFE except SeaCache (7 NFE). SpecKCast achieves $5.61\times$ speedup while avoiding the strong color and boundary artifacts of HiCache and TaylorSeer. TeaCache ($5.64\times$) and SeaCache ($6.05\times$) are faster, but TeaCache softens the panda's detail and SeaCache changes the scene more substantially.}
  \label{fig:app_wan_panda}

  \vspace{1.5em}

  \includegraphics[width=\linewidth]{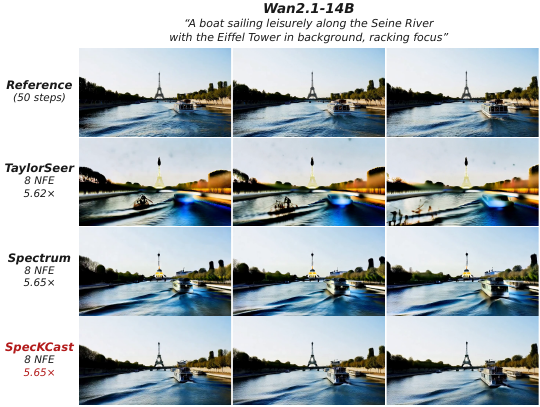}
  \caption{\textbf{Wan2.1-14B: a boat on the Seine with the Eiffel Tower in the background.} All three accelerated methods use 8 NFE, with speedups of $5.62\times$ (TaylorSeer) and $5.65\times$ (Spectrum and SpecKCast). SpecKCast more closely preserves the tower's silhouette and scene layout across the displayed frames.}
  \label{fig:app_wan_boat}
\end{figure}

\clearpage
\section{Additional Spatial-Frequency Trajectory Analysis}
\label{app:trajectories}

We extend the single-generation FLUX.1-dev trajectories in \Cref{fig:fig1_velocity} in three ways: joint views computed over 12 prompts, individual generations for six of the prompts, and the same analysis for Qwen-Image. Both collections use the same 12 prompts, with one generation per prompt and model. \Cref{fig:app_flux_pooled} shows the joint FLUX.1-dev trajectories in the PC1--PC2 plane and as scores against denoising step, and \Cref{fig:app_qwen_pooled} shows all three joint views for Qwen-Image. The remaining figures show individual generations for six of the prompts. Together, these visualizations examine how temporal evolution differs across spatial-frequency bands and how the patterns vary across prompts and models.

\subsection{Measurement Setup}
\label{app:trajectory_measurements}

For both models, each generation uses a resolution of $1024\times1024$, 50 denoising steps indexed from 0 to 49, and full transformer computation at every step. We use zero-based prompt IDs: prompt $p\in\{0,\ldots,11\}$ uses generation seed $p$. The texts of prompts 0--5, which are shown individually, are given in the figure captions.

In both models, the recorded velocity is unpacked to $Y_j\in\mathbb{R}^{16\times128\times128}$, and we apply an orthonormal two-dimensional DCT-II independently to each channel. For spatial-frequency indices $(a,b)$, the normalized radius is
\begin{equation}
r(a,b)=\frac{\sqrt{(a/128)^2+(b/128)^2}}{\sqrt{2}},
\qquad a,b\in\{0,\ldots,127\}.
\label{eq:app_qwen_radius}
\end{equation}
The seven radial bands have edges $[0,1/64,1/32,1/16,1/8,1/4,1/2,1]$. Each panel set visualizes 512 fixed sampled entries of the full velocity, all 128 coefficients of $B_0$, and 512 fixed sampled coefficients each from $B_3$ and $B_6$. Coefficient sampling uses seed 1234, and the sampled indices are shared across steps and prompts. The full-velocity panel summarizes sampled entries of the spatial-domain output, while the band panels summarize DCT coefficients. The joint analysis combines the measurements from all 12 generations, while each individual analysis uses only its own generation.

\paragraph{FLUX.1-dev.}
The FLUX collection uses guidance scale 3.5, and we record the velocity produced by the transformer at each step.

\paragraph{Qwen-Image.}
We record the velocity passed to the sampler after classifier-free guidance at scale 4 and the pipeline's norm rescaling. For conditional and unconditional predictions $c_{j,t}$ and $u_{j,t}$ at packed token $t$, the recorded output is
\begin{equation}
\widetilde v_{j,t}=u_{j,t}+4(c_{j,t}-u_{j,t}),
\qquad
v_{j,t}=\widetilde v_{j,t}\,
\frac{\lVert c_{j,t}\rVert_2}{\lVert\widetilde v_{j,t}\rVert_2},
\label{eq:app_qwen_guidance}
\end{equation}
where the negative prompt is empty and each norm is taken over the packed token's 64 channels.

\subsection{PCA Construction and Display}
\label{app:trajectory_pca}

Both models use the same procedure for joint views and normalization, applied independently to each model. For each prompt and panel, we arrange the recorded values into a coefficient-by-step matrix and divide it by its Frobenius norm. For a joint view, the 12 normalized matrices are concatenated along the coefficient axis and transposed, giving a matrix with 50 observations and $12d$ features, where $d$ is the number of retained entries in that panel. We center each feature across steps and project onto its first two principal components. Thus, each joint curve summarizes temporal variation across all 12 generations. For the individual views, the same procedure is applied to each prompt separately, giving a distinct PCA basis for each prompt and panel.

Within each panel, both score sequences are divided by their joint maximum absolute value. This shared scaling preserves the relative geometry of the two displayed components. For both models, the PCA is computed by singular value decomposition, PC1 is oriented so that its final score is at least its initial score, and PC2 is oriented so that its score at step 25 is nonnegative. Within each joint figure, all rows use exactly the same normalized scores, as do the paired three-dimensional and planar views for each individual prompt. Planar views use equal axis scaling and shared square limits. Circles mark measured states, blue changes from dark to light with denoising progress, and stars mark the final state (PC1 in the score-versus-step view). Straight segments connect successive measurements without smoothing.

\subsection{Temporal Patterns Across Prompts}
\label{app:trajectory_interpretation}

\paragraph{FLUX.1-dev.}
\Cref{fig:app_flux_pooled} shows that the pattern of the single generation in \Cref{fig:fig1_velocity} holds across all 12 prompts. In $B_0$, the displayed scores change sharply during the first few steps and remain near their final values over most of the schedule. The middle band $B_3$ evolves over a broader interval. The high band $B_6$ changes slowly at first, with larger changes near the end. The planar projections show the path of the measured states, while the score-versus-step curves show when each component evolves.

\paragraph{Qwen-Image.}
In \Cref{fig:app_qwen_pooled}, $B_0$ also undergoes a sharp initial transient and then changes little in the displayed components. The middle band $B_3$ evolves over a broader portion of the schedule, while $B_6$ exhibits an early transient followed by a pronounced change near the end. Across both models, the different timing across bands supports the motivation in \Cref{sec:freq_decomp} for separate temporal models and band-specific forecasting coordinates.

Each panel has its own PCA basis and normalization, so absolute distances, directions, and amplitudes cannot be compared across models, bands, prompts, or joint and individual views. A nearly vertical trajectory in the three-dimensional plots indicates little change in PC1 and PC2; variation can still occur in the omitted components. These plots therefore provide qualitative evidence about temporal structure in the retained subspaces, rather than a quantitative measure of full-velocity motion or forecasting error.

\subsection{Individual Prompt Trajectories}
\label{app:trajectory_individual}

\Cref{fig:app_flux_000_001,fig:app_flux_002_003,fig:app_flux_004_005} show individual FLUX.1-dev generations for prompts 0--5, and \Cref{fig:app_qwen_000_001,fig:app_qwen_002_003,fig:app_qwen_004_005} show the same prompts for Qwen-Image, with two prompts per figure. For each prompt, the upper row displays the trajectory with denoising step increasing downward, and the lower row displays the identical normalized PC pairs in the plane. Columns show full velocity, $B_0$, $B_3$, and $B_6$. In the planar views, numbers next to points mark denoising steps: steps 0, 25, and 49 for the full velocity, and steps 0 and 49 together with step 1 for $B_0$, step 10 for $B_3$, and step 40 for $B_6$, marking where each band changes most. These separate fits show differences between generations that are averaged out in the joint views. Their coordinates are specific to each model, prompt, and panel.

\begin{figure}[p]
  \centering
  \includegraphics[width=\linewidth]{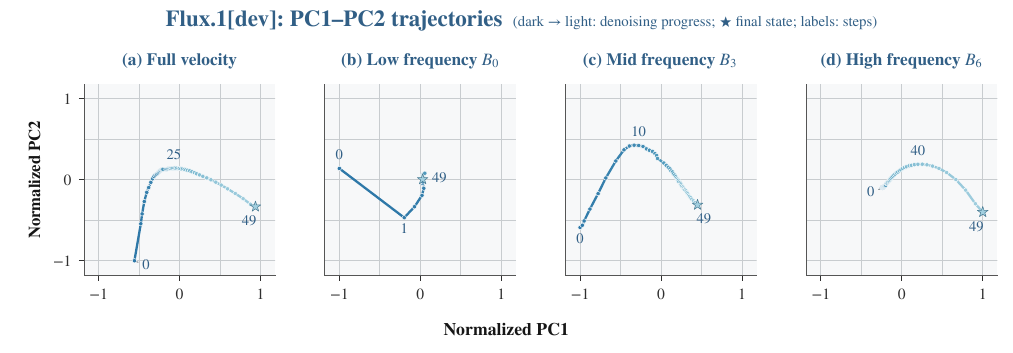}
  \par\medskip
  \includegraphics[width=\linewidth]{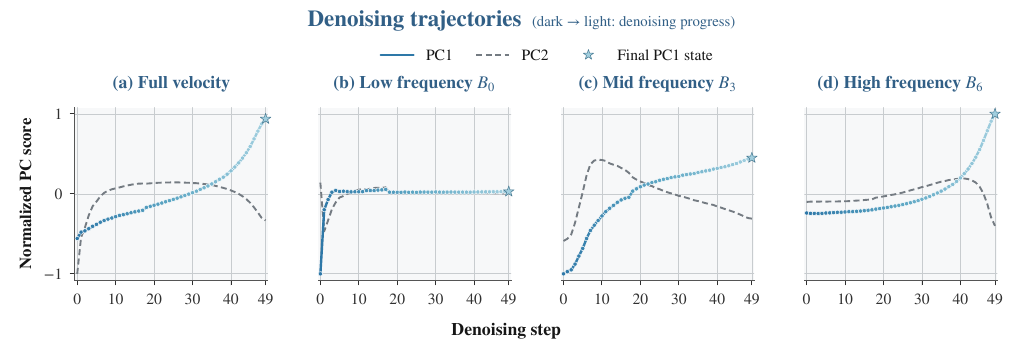}
  \caption{\textbf{Joint FLUX.1-dev trajectories across 12 prompts.} Columns show the sampled full velocity and DCT bands $B_0$, $B_3$, and $B_6$. Both rows use the same normalized PCA scores, computed jointly over 12 prompts with 50 measured velocities per prompt, guidance scale 3.5, and full transformer computation. Unlike \Cref{fig:fig1_velocity}, which shows a single generation, the PCA here is fitted to all 12 generations together. \textbf{Top:} PC1--PC2 projections with equal axis scaling; labels identify selected steps. \textbf{Bottom:} PC1 (solid blue) and PC2 (dashed gray) against step index. Circles mark measurements, dark to light blue indicates denoising progress, and stars mark the final state (PC1 in the bottom row). Each column has its own PCA basis and scale, constructed as described in \Cref{app:trajectory_pca}.}
  \label{fig:app_flux_pooled}
\end{figure}

\begin{figure}[p]
  \centering
  \includegraphics[width=\linewidth]{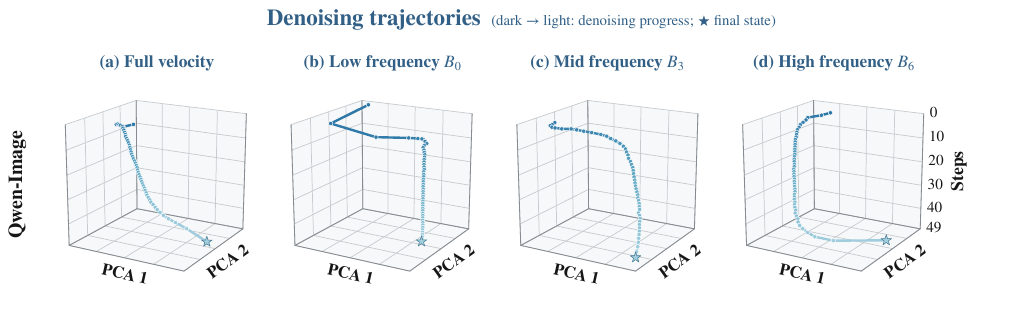}
  \par\smallskip
  \includegraphics[width=\linewidth]{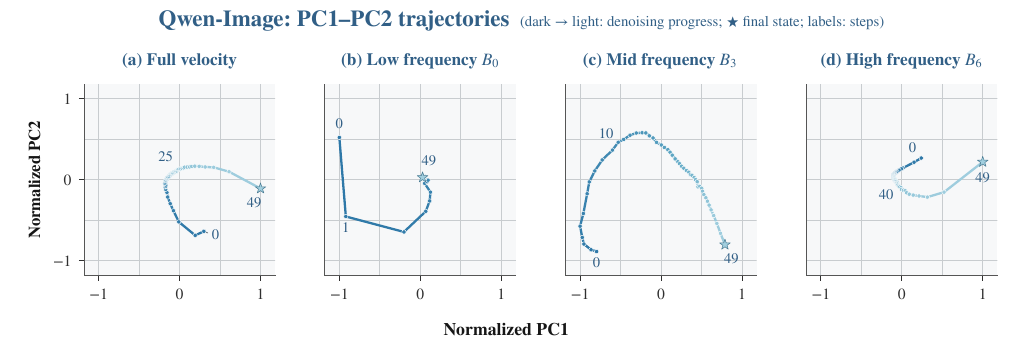}
  \par\smallskip
  \includegraphics[width=\linewidth]{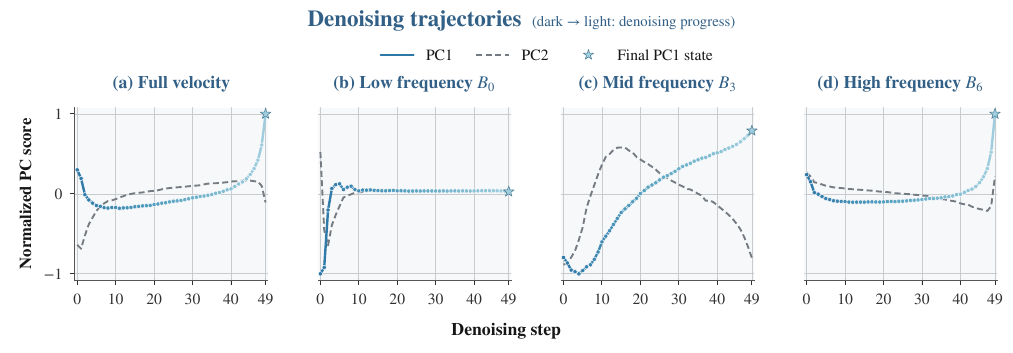}
  \caption{\textbf{Spatial-frequency trajectories of Qwen-Image across 12 prompts.} Columns show the sampled full velocity and DCT bands $B_0$, $B_3$, and $B_6$. All rows use the same normalized PCA scores, computed jointly over 12 prompts with 50 velocities passed to the sampler per prompt, guidance scale 4, and full transformer computation. \textbf{Top:} Denoising step increases downward. \textbf{Middle:} The same PC1--PC2 pairs with equal axis scaling; labels identify selected steps. \textbf{Bottom:} PC1 (solid blue) and PC2 (dashed gray) against step index. The low band changes mainly at the start, the middle band evolves over a broader interval, and the high band changes sharply near the end after a small early transient. Bases and scales are fitted independently for each column; see \Cref{app:trajectory_pca} for the construction.}
  \label{fig:app_qwen_pooled}
\end{figure}

\begin{figure}[p]
  \centering
  {\small\textbf{Prompt 000 (seed 0)}\par}
  \includegraphics[width=0.95\linewidth]{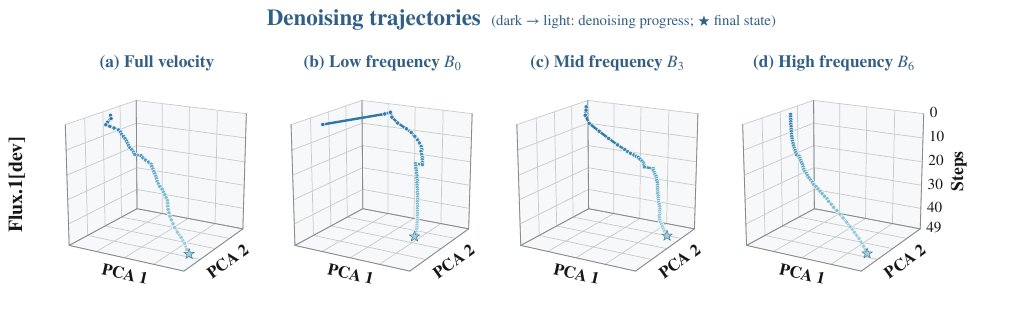}
  \par
  \includegraphics[width=0.95\linewidth]{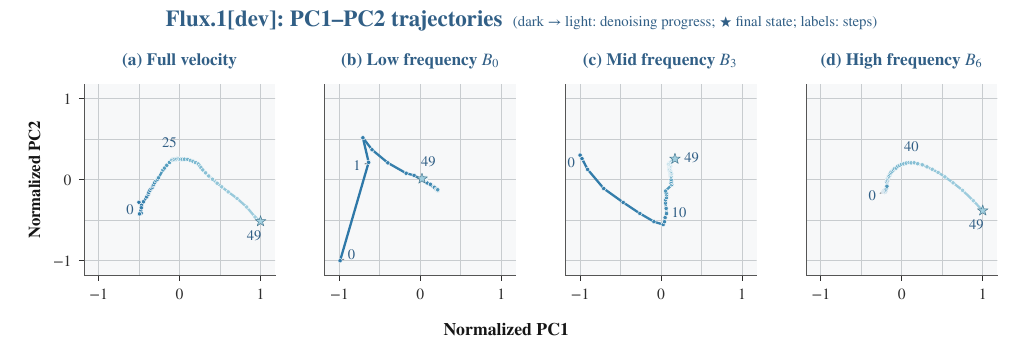}
  \par\medskip
  {\small\textbf{Prompt 001 (seed 1)}\par}
  \includegraphics[width=0.95\linewidth]{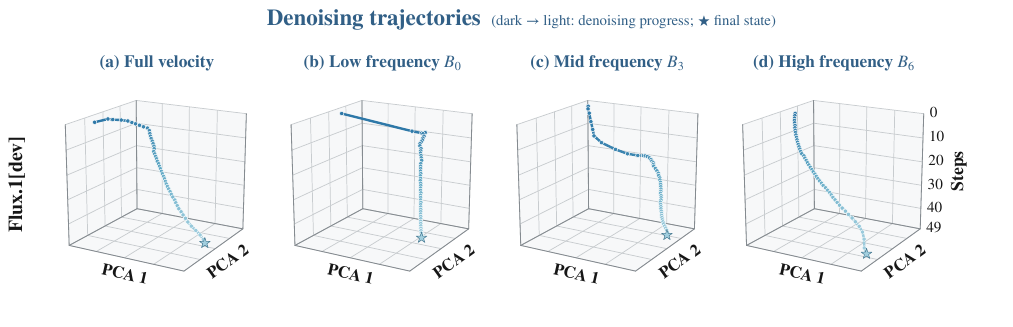}
  \par
  \includegraphics[width=0.95\linewidth]{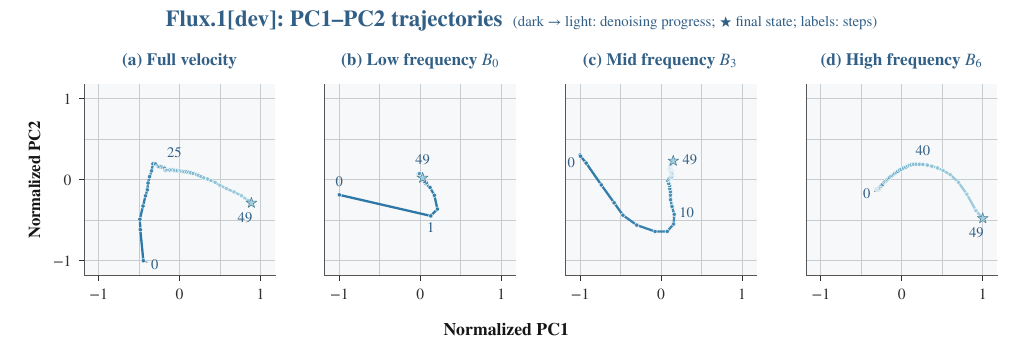}
  \caption{\textbf{Individual FLUX.1-dev trajectories: prompts 000 and 001.} Top two rows: ``a cat holding a red apple'' (seed 0). Bottom two rows: ``a photograph of an astronaut riding a horse on the moon'' (seed 1). For each prompt, the same 50 states are shown in 3D with the denoising step as the vertical axis and in the PC1--PC2 plane. Columns show full velocity, $B_0$, $B_3$, and $B_6$; dark to light blue indicates denoising progress and stars mark the final state. Each panel uses its own PCA basis and scaling, so positions and sizes cannot be compared across panels or prompts.}
  \label{fig:app_flux_000_001}
\end{figure}

\begin{figure}[p]
  \centering
  {\small\textbf{Prompt 002 (seed 2)}\par}
  \includegraphics[width=0.95\linewidth]{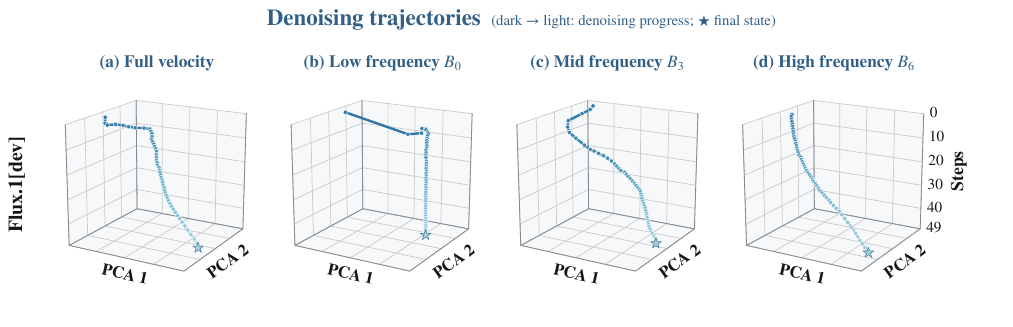}
  \par
  \includegraphics[width=0.95\linewidth]{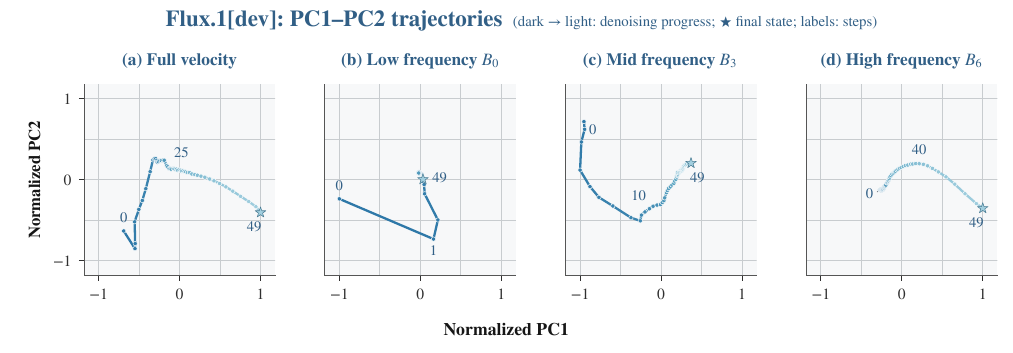}
  \par\medskip
  {\small\textbf{Prompt 003 (seed 3)}\par}
  \includegraphics[width=0.95\linewidth]{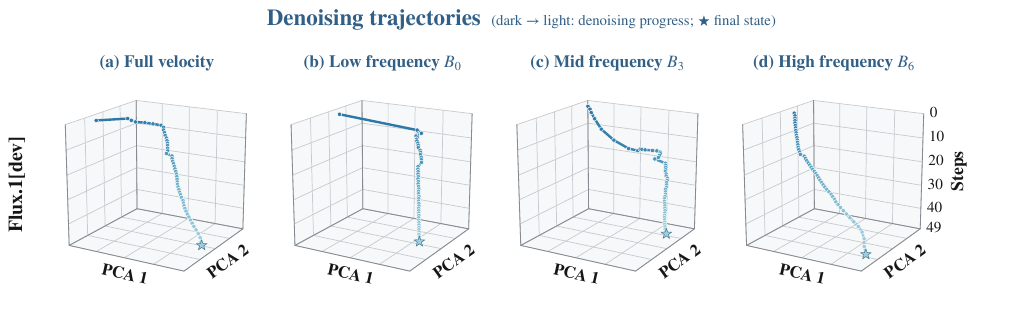}
  \par
  \includegraphics[width=0.95\linewidth]{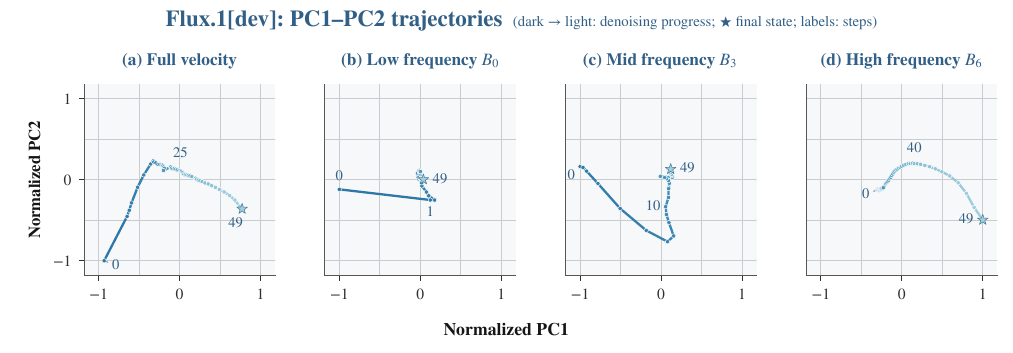}
  \caption{\textbf{Individual FLUX.1-dev trajectories: prompts 002 and 003.} Top two rows: ``a bowl of ramen with chopsticks, steam rising, cinematic lighting'' (seed 2). Bottom two rows: ``a red panda reading a book in a cozy library'' (seed 3). For each prompt, the same 50 states are shown in 3D with the denoising step as the vertical axis and in the PC1--PC2 plane. Columns show full velocity, $B_0$, $B_3$, and $B_6$; dark to light blue indicates denoising progress and stars mark the final state. Each panel uses its own PCA basis and scaling, so positions and sizes cannot be compared across panels or prompts.}
  \label{fig:app_flux_002_003}
\end{figure}

\begin{figure}[p]
  \centering
  {\small\textbf{Prompt 004 (seed 4)}\par}
  \includegraphics[width=0.95\linewidth]{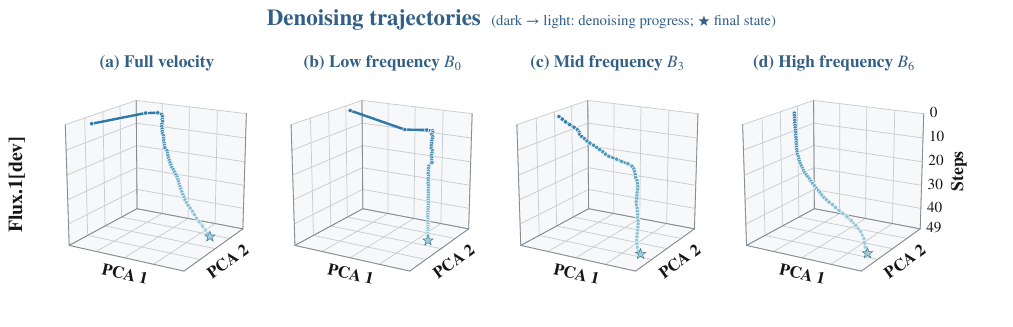}
  \par
  \includegraphics[width=0.95\linewidth]{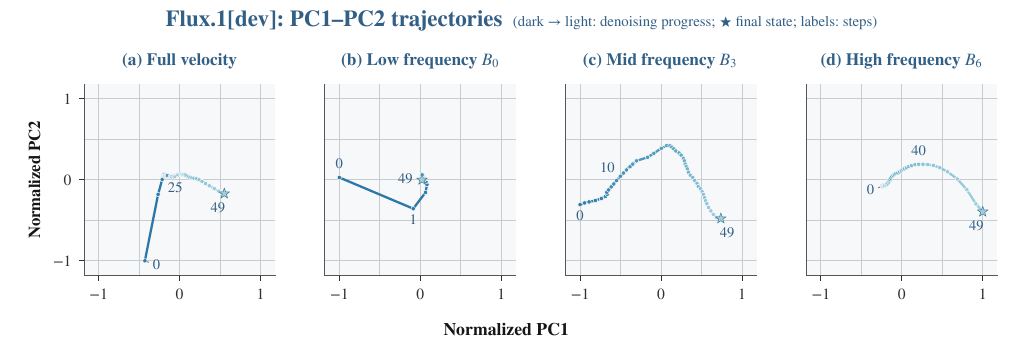}
  \par\medskip
  {\small\textbf{Prompt 005 (seed 5)}\par}
  \includegraphics[width=0.95\linewidth]{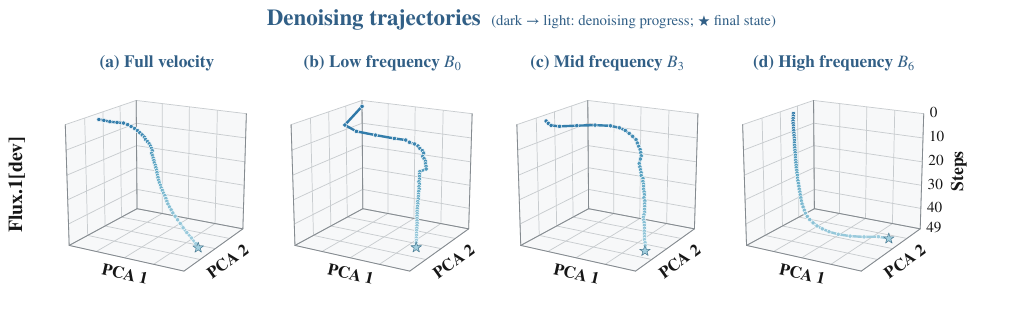}
  \par
  \includegraphics[width=0.95\linewidth]{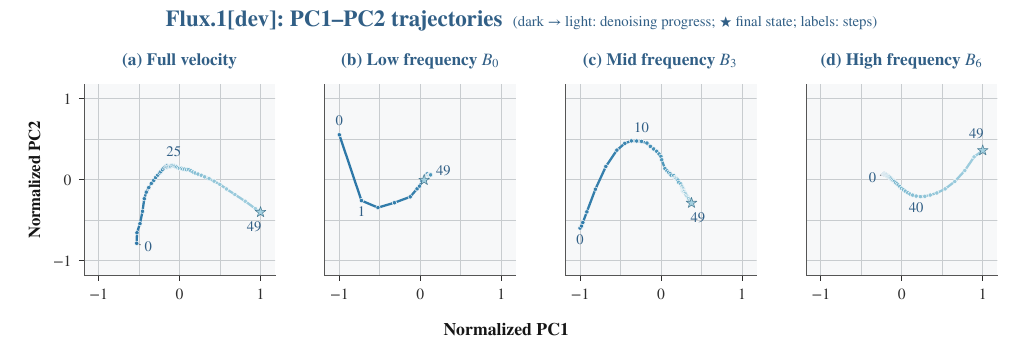}
  \caption{\textbf{Individual FLUX.1-dev trajectories: prompts 004 and 005.} Top two rows: ``a futuristic city skyline at sunset with flying cars'' (seed 4). Bottom two rows: ``a close-up portrait of an old fisherman with a weathered face'' (seed 5). For each prompt, the same 50 states are shown in 3D with the denoising step as the vertical axis and in the PC1--PC2 plane. Columns show full velocity, $B_0$, $B_3$, and $B_6$; dark to light blue indicates denoising progress and stars mark the final state. Each panel uses its own PCA basis and scaling, so positions and sizes cannot be compared across panels or prompts.}
  \label{fig:app_flux_004_005}
\end{figure}

\begin{figure}[p]
  \centering
  {\small\textbf{Prompt 000 (seed 0)}\par}
  \includegraphics[width=0.95\linewidth]{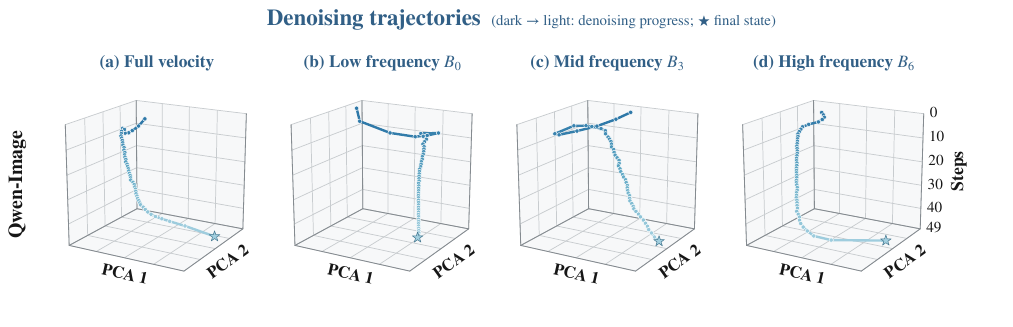}
  \par
  \includegraphics[width=0.95\linewidth]{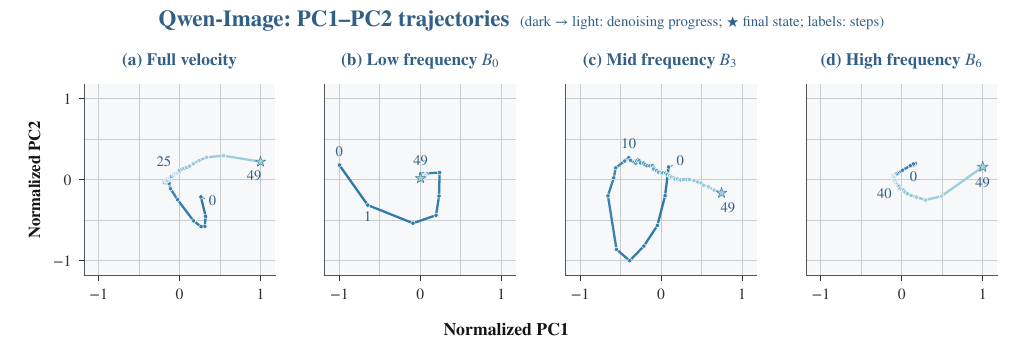}
  \par\medskip
  {\small\textbf{Prompt 001 (seed 1)}\par}
  \includegraphics[width=0.95\linewidth]{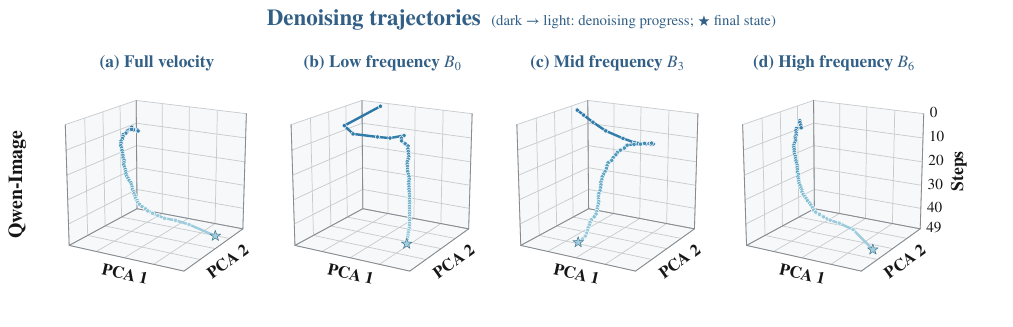}
  \par
  \includegraphics[width=0.95\linewidth]{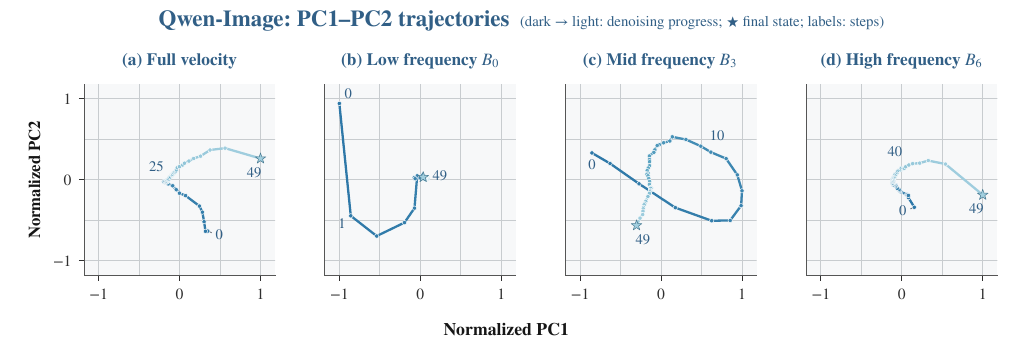}
  \caption{\textbf{Individual Qwen-Image trajectories: prompts 000 and 001.} Top two rows: ``a cat holding a red apple'' (seed 0). Bottom two rows: ``a photograph of an astronaut riding a horse on the moon'' (seed 1). For each prompt, the same 50 states are shown in 3D with the denoising step as the vertical axis and in the PC1--PC2 plane. Columns show full velocity, $B_0$, $B_3$, and $B_6$; dark to light blue indicates denoising progress and stars mark the final state. Each panel uses its own PCA basis and scaling, so positions and sizes cannot be compared across panels or prompts.}
  \label{fig:app_qwen_000_001}
\end{figure}

\begin{figure}[p]
  \centering
  {\small\textbf{Prompt 002 (seed 2)}\par}
  \includegraphics[width=0.95\linewidth]{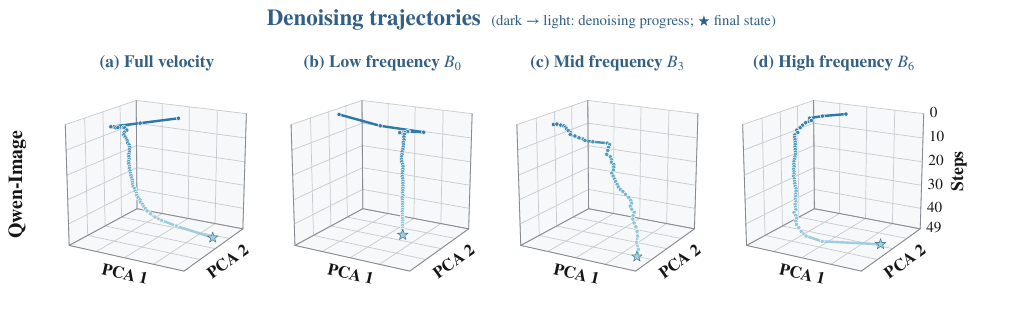}
  \par
  \includegraphics[width=0.95\linewidth]{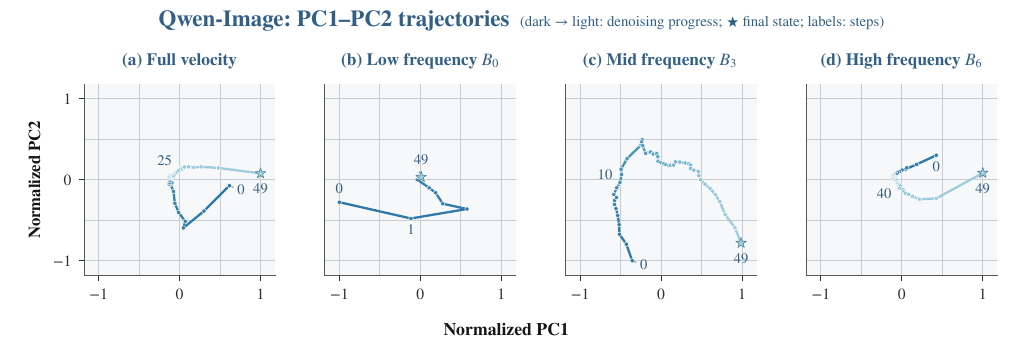}
  \par\medskip
  {\small\textbf{Prompt 003 (seed 3)}\par}
  \includegraphics[width=0.95\linewidth]{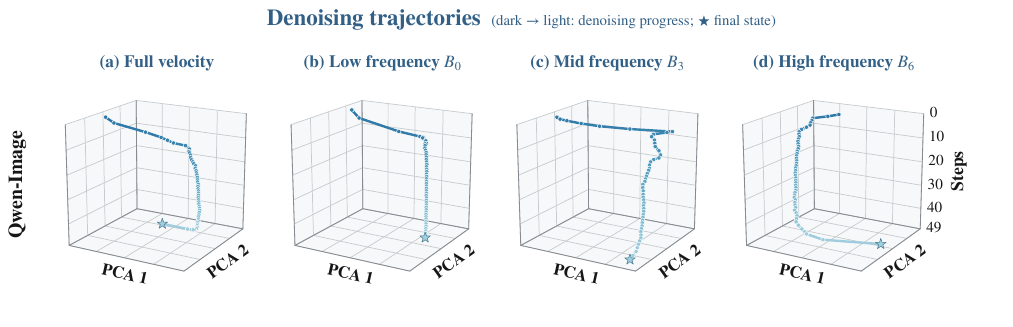}
  \par
  \includegraphics[width=0.95\linewidth]{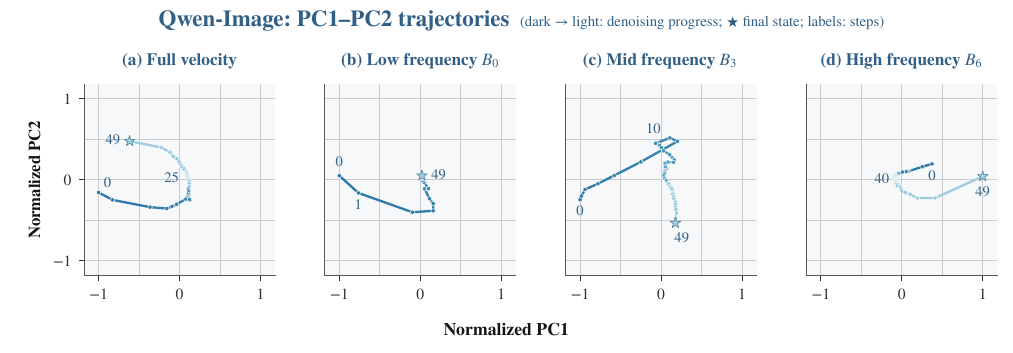}
  \caption{\textbf{Individual Qwen-Image trajectories: prompts 002 and 003.} Top two rows: ``a bowl of ramen with chopsticks, steam rising, cinematic lighting'' (seed 2). Bottom two rows: ``a red panda reading a book in a cozy library'' (seed 3). For each prompt, the same 50 states are shown in 3D with the denoising step as the vertical axis and in the PC1--PC2 plane. Columns show full velocity, $B_0$, $B_3$, and $B_6$; dark to light blue indicates denoising progress and stars mark the final state. Each panel uses its own PCA basis and scaling, so positions and sizes cannot be compared across panels or prompts.}
  \label{fig:app_qwen_002_003}
\end{figure}

\begin{figure}[p]
  \centering
  {\small\textbf{Prompt 004 (seed 4)}\par}
  \includegraphics[width=0.95\linewidth]{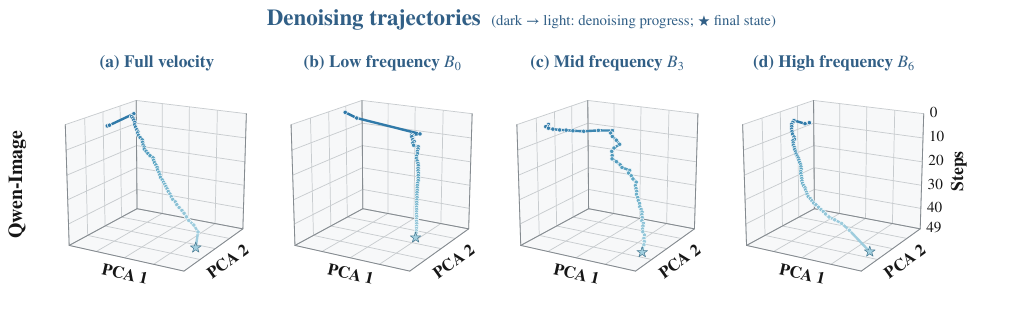}
  \par
  \includegraphics[width=0.95\linewidth]{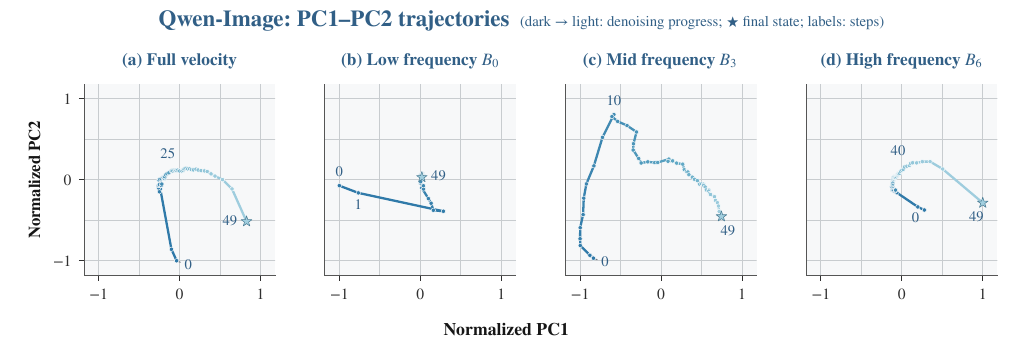}
  \par\medskip
  {\small\textbf{Prompt 005 (seed 5)}\par}
  \includegraphics[width=0.95\linewidth]{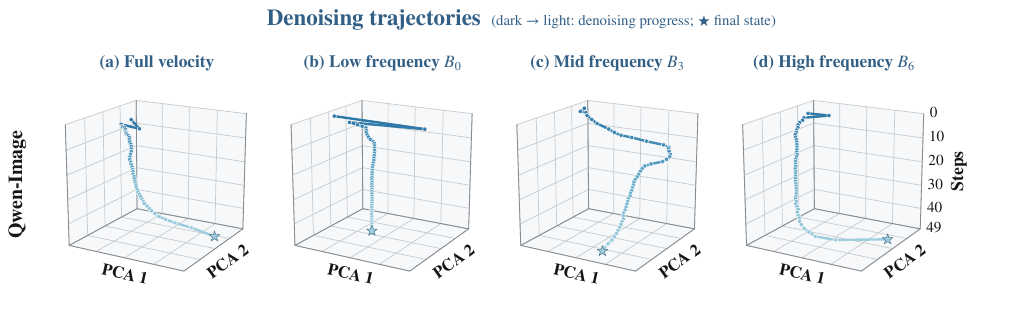}
  \par
  \includegraphics[width=0.95\linewidth]{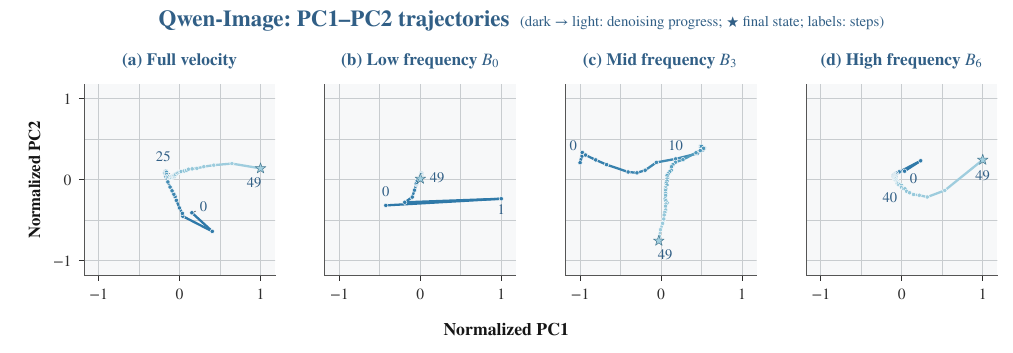}
  \caption{\textbf{Individual Qwen-Image trajectories: prompts 004 and 005.} Top two rows: ``a futuristic city skyline at sunset with flying cars'' (seed 4). Bottom two rows: ``a close-up portrait of an old fisherman with a weathered face'' (seed 5). For each prompt, the same 50 states are shown in 3D with the denoising step as the vertical axis and in the PC1--PC2 plane. Columns show full velocity, $B_0$, $B_3$, and $B_6$; dark to light blue indicates denoising progress and stars mark the final state. Each panel uses its own PCA basis and scaling, so positions and sizes cannot be compared across panels or prompts.}
  \label{fig:app_qwen_004_005}
\end{figure}

\end{document}